\def\paperID{6225}
\def\confName{CVPR}
\def\confYear{2025}

\def\paperTitle{GBlobs: Explicit Local Structure via Gaussian Blobs for Improved Cross-Domain LiDAR-based 3D Object Detection}

\def\authorBlock{
    Du\v{s}an Mali\'c$^{1,2}$ \qquad
    Christian Fruhwirth-Reisinger$^{1,2}$ \qquad
    Samuel Schulter$^{3,\dag}$ \qquad
    Horst Possegger$^{1,2}$ \\
    $^1$Christian Doppler Laboratory for Embedded Machine Learning \\
    $^2$Institute of Visual Computing, Graz University of Technology\\
    $^3$Amazon \\
    {\tt\small \{dusan.malic, reisinger, possegger\}@tugraz.at}
}

\newif\ifreview 
\newif\ifarxiv \newcommand{\arxiv}{\arxivtrue}
\newif\ifcamera 
\newif\ifrebuttal 

\arxiv %

\pdfoutput=1
\documentclass[10pt,twocolumn,letterpaper]{article}
\ifreview \usepackage[review]{cvpr} \fi
\ifarxiv \usepackage[pagenumbers]{cvpr} \fi
\ifrebuttal \usepackage[rebuttal]{cvpr} \fi
\ifcamera \usepackage{cvpr} \fi

\usepackage{graphicx}	
\usepackage{amsmath}	
\usepackage{amssymb}	
\usepackage{booktabs}
\usepackage{times}
\usepackage{microtype}
\usepackage{epsfig}
\usepackage{caption}
\usepackage{float}
\usepackage{placeins}
\usepackage{color, colortbl}
\usepackage{stfloats}
\usepackage{enumitem}
\usepackage{tabularx}
\usepackage{xstring}
\usepackage{multirow}
\usepackage{xspace}
\usepackage{url}
\usepackage{subcaption}
\usepackage{xcolor}
\usepackage[hang,flushmargin]{footmisc}

\usepackage{graphicx}
\usepackage{amsmath}
\usepackage{amssymb}
\usepackage{booktabs}
\usepackage{multirow}
\usepackage{bm}
\usepackage{algorithm}
\usepackage{listings}
\usepackage{caption}
\usepackage{subcaption}
\usepackage{csquotes}
\usepackage{mathabx}
\usepackage{soul}
\usepackage{multirow}
\usepackage{pifont}

\ifcamera \usepackage[accsupp]{axessibility} \fi

\newcommand{\nbf}[1]{{\noindent \textbf{#1.}}}

\ifarxiv  \fi

\definecolor{violetish}{RGB}{120, 20, 180}
\newcommand{\R}[1]{{%
    \textbf{%
        \ifstrequal{#1}{1}{\textcolor{violetish}{Quhi}}{%
        \ifstrequal{#1}{2}{\textcolor{teal}{iEde}}{%
        \ifstrequal{#1}{3}{\textcolor{cyan}{RLiZ}}{%
        \ifstrequal{#1}{4}{\textcolor{magenta}{R#1}}{%
                           \textcolor{red}{R#1}%
        }}}}%
    }%
}}

\newcommand{\sota}{state-of-the-art\xspace}

\newcommand\blfootnote[1]{%
  \begingroup
  \renewcommand\thefootnote{}\footnote{#1}%
  \addtocounter{footnote}{-1}%
  \endgroup
}

\newcommand{\WtoN}{Waymo$\rightarrow$nuScenes\xspace}
\newcommand{\WtoK}{Waymo$\rightarrow$KITTI\xspace}

\newcommand{\NtoK}{nuScenes$\rightarrow$KITTI\xspace}
\newcommand{\KtoW}{KITTI$\rightarrow$Waymo\xspace}

\newcommand{\threedvf}{3D-VF\xspace}

\newcommand{\bestresult}[1]{\text{\textbf{#1}}}

\newcommand{\nbfnodot}[1]{{\noindent \textbf{#1}}}

\newcommand{\ourmethod}{GBlobs\xspace}

\usepackage{tikz}
\usepackage{pgfplots}
\pgfplotsset{compat=1.17}

\usepackage{xr-hyper}

\makeatletter
\newcommand*{\addFileDependency}[1]{
  \typeout{(#1)}
  \@addtofilelist{#1}
  \IfFileExists{#1}{}{\typeout{No file #1.}}
}

\makeatother
\newcommand*{\myexternaldocument}[1]{
    \externaldocument{#1}
    \addFileDependency{#1.tex}
    \addFileDependency{#1.aux}
}

\definecolor{cvprblue}{rgb}{0.21,0.49,0.74}
\usepackage[pagebackref,breaklinks,colorlinks,allcolors=cvprblue]{hyperref}
\usepackage[capitalize]{cleveref}
\crefname{section}{Sec.}{Secs.}
\crefname{table}{Table}{Tables}
\crefname{figure}{Fig.}{Figs.}

\ifarxiv \crefname{appendix}{App.}{Apps.}
\else \crefname{appendix}{Suppl.}{Suppls.} \fi

\unless\ifarxiv \myexternaldocument{_supplementary} \fi

\begin{document}
\title{\paperTitle}
\author{\authorBlock}
\maketitle

\blfootnote{$^\dag$Work conducted prior to joining Amazon}

\begin{abstract}
    LiDAR-based 3D detectors need large datasets for training, yet they struggle to generalize to novel domains.
    Domain Generalization (DG) aims to mitigate this by training detectors that are invariant to such domain shifts.
    Current DG approaches exclusively rely on \textbf{global} geometric features (point cloud Cartesian coordinates) as input features.
    Over-reliance on these global geometric features can, however, cause 3D detectors to prioritize object location and absolute position, resulting in poor cross-domain performance.
    To mitigate this, we propose to exploit explicit \textbf{local} point cloud structure for DG, in particular by encoding point cloud neighborhoods with Gaussian blobs, \textbf{\ourmethod}.
    Our proposed formulation is highly efficient and requires no additional parameters.
    Without any bells and whistles, simply by integrating \textbf{\ourmethod} in existing detectors, we beat the current \sota in challenging single-source DG benchmarks by over $21$ mAP (\WtoK), $13$ mAP (\KtoW), and $12$ mAP (\NtoK), without sacrificing in-domain performance.
    Additionally, \ourmethod demonstrate exceptional performance in multi-source DG, surpassing the current \sota by $17$, $12$, and $5$ mAP on Waymo, KITTI, and ONCE, respectively.
\end{abstract}

\section{Introduction}
\label{sec:intro}

LiDAR-based 3D detection models \cite{chen2023voxelnext,yin2021center,shi2023pv,shi2019pointrcnn,yan2018second} predominantly rely on \emph{global} input features, \ie representing points by their Cartesian coordinates.
Only a small subset of research \emph{additionally} incorporates \emph{local} point cloud information, like relative distance to the voxel center~\cite{deng2021voxel,wang2023dsvt}, LiDAR intensity~\cite{naich2024lidar}, surface normals~\cite{zhao2021integration,miao20213d} or handcrafted local descriptors~\cite{li2024trail} to enhance model performance.
The convenience of global input features and their strong in-domain performance established this representation as the default choice for most \sota detection models.
However, such models are highly sensitive to rigid transformations~\cite{li2024trail,wang2024ride}, being more biased toward object position than local features like shape or appearance.
Therefore, they generalize poorly to novel, unseen domains~\cite{3dda:st3d,yang2022st3d++,eskandar2024empirical,wang2020trainingermany}.

\begin{figure}
     \centering
     \begin{subfigure}{\columnwidth}
         \centering
         \includegraphics[width=0.9\textwidth]{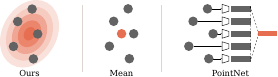}
         \caption{
           Our proposed encoder estimates Gaussian blobs from local neighborhoods, providing a more expressive representation of local geometry compared to PointNet~\cite{Qi_2017_CVPR} and the more commonly used mean encoding.
         }
         \label{fig:motivation different vfe}
     \end{subfigure}
     \begin{subfigure}{0.48\columnwidth}
         \centering
         \includegraphics[width=\textwidth]{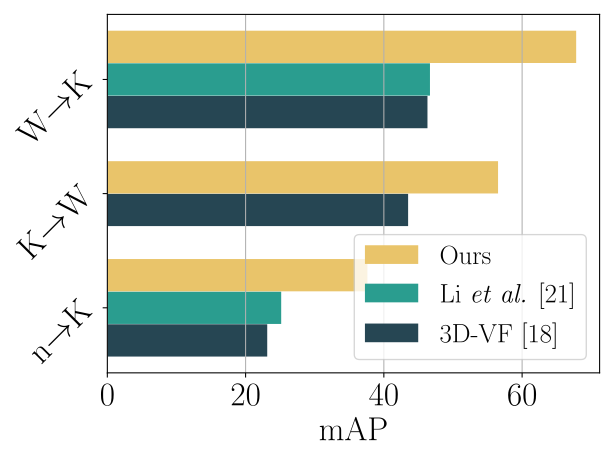}
         \caption{
           Comparison with \sota single-source DG methods.
           Benchmark results for \WtoK (W$\rightarrow$K), \KtoW (K$\rightarrow$W), and \NtoK (n$\rightarrow$K).
         }
         \label{fig:motivation barplot single source}
     \end{subfigure}
     \hfill
     \begin{subfigure}{0.48\columnwidth}
         \centering
         \includegraphics[width=\textwidth]{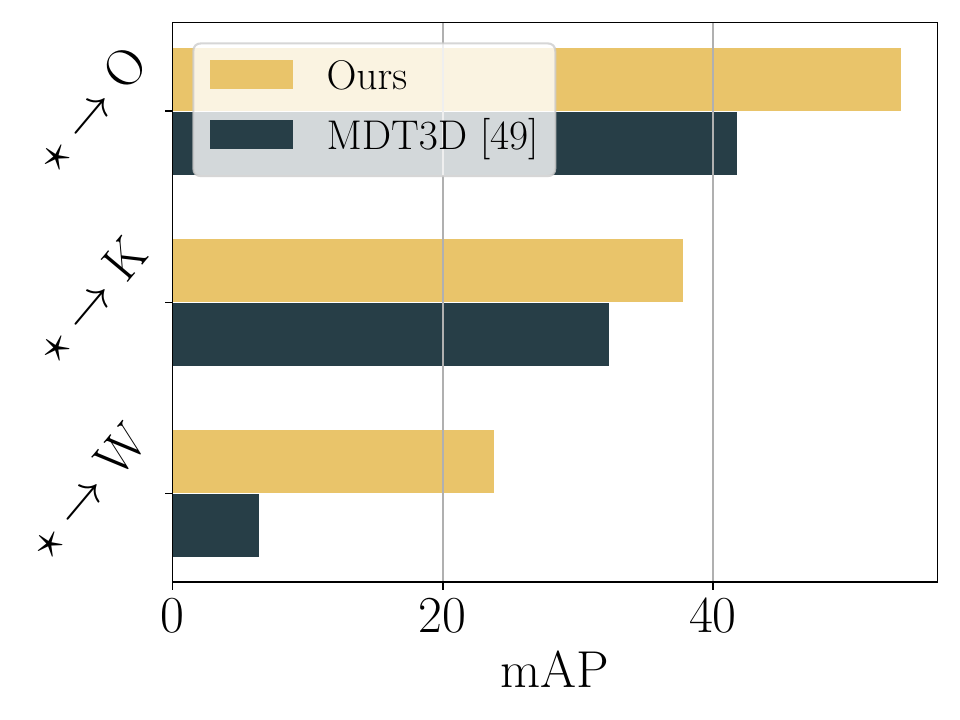}
         \caption{
           Comparison with \sota multi-sorce DG methods.
           Benchmark results from training on all ($*$), excluding the target dataset, and evaluating it on ONCE (O), KITTI (K) and Waymo (W).
         }
         \label{fig:motivation barplot multi source}
     \end{subfigure}
     \caption{
       We propose a novel encoding scheme for local neighborhood in point clouds, using a Gaussian blob instead of the commonly employed mean or PointNet encoding (a).
       Our simple yet effective approach significantly outperforms \sota methods in both single- and multi-source domain generalization does not require any additional parameters, spatial training procedures, or hyperparameter optimization (b, c).
    }
     \label{fig:motivation bar plots}
\end{figure}

In this paper, we demonstrate that local point cloud geometry can significantly improve model generalization.
In particular, we introduce a new method for representing local neighborhoods as Gaussian blobs (see \cref{fig:motivation different vfe}), defined by their means and covariance matrices.
Our representation effectively decouples object location from model encoding, enabling the model to focus on learning local shape and appearance.
Unlike existing local descriptors based on relative distances~\cite{deng2021voxel,wang2023dsvt} or surface normals~\cite{hoppe1992surface}, our formulation is more descriptive due to the covariance modeling and eliminates ambiguities arising from surface orientation.
Moreover, our method is inherently invariant to point permutations (\ie a neighborhood's covariance matrix remains the same regardless of the point order), which eliminates the need for point cloud sorting~\cite{ran2022surface,li2024trail,li2024rapid} or channel-wise max pooling and concatenation~\cite{Qi_2017_CVPR} (in \cref{fig:motivation different vfe}).
To demonstrate the effectiveness of our approach, we evaluate it on challenging cross-domain LiDAR 3D object detection benchmarks, highlighting the significant improvements on \emph{domain generalization} (DG) for a wide range of 3D detectors.
While previous methods often require specialized data augmentation~\cite{lehner20223d,soum2023mdt3d}, auxiliary tasks~\cite{li2023domainne2021}, or domain-invariant feature learning~\cite{wu2023towards}, which can introduce additional complexity and limitations, our module is simple, effective, universally applicable to any dataset and model, and does not degrade the in-domain performance.

In the single-domain generalization benchmarks, as shown in \cref{fig:motivation barplot single source}, we achieve significant improvements over the \sota, with over $21$ mAP in \WtoK (W$\rightarrow$K), $13$ mAP in \KtoW (K$\rightarrow$W), and $12$ mAP in \NtoK (n$\rightarrow$K).
In our multi-domain DG experiments in \cref{fig:motivation barplot multi source}, where we trained on all datasets ($*$) except the target dataset, we consistently observed significant improvements over the \sota, with gains of over $17$, $12$, and $5$ mAP on Waymo, KITTI, and ONCE, respectively.
In summary, our main contributions are:
\begin{itemize}
  \item A novel, permutation-invariant LiDAR point cloud representation using Gaussian blobs, offering descriptive features, efficient computation, and no additional model parameters.
  \item Comprehensive evaluations on both single- and multi-source domain generalization benchmarks and in-depth ablation studies.
  \item Publicly accessible code and trained models at \url{https://github.com/malicd/GBlobs}

\end{itemize}

\section{Related Work}
\label{sec:relwork}

\nbfnodot{LiDAR-based 3D object detection}
can be broadly classified by their input representations.
Methods such as~\cite{yang20203dssd, shi2019pointrcnn, qi2018frustum, wu2024ptv3} operate directly on unordered point cloud data.
To address the computational challenges of processing raw point clouds,~\cite{meyer2019lasernet} introduces a LiDAR range view representation and leverages efficient 2D convolutions.
Similarly, pillar-based networks~\cite{lang2019pointpillars, li2023pillarnext, fan2022embracing, sun2022swformer} encode 3D data into a sparse 2D grid where each grid cell (pillar) spans the entire height axis.
To address the information loss inherent in such dimension reduction, voxel-based methods~\cite{deng2021voxel, chen2023voxelnext, yin2021center, yan2018second, wang2023dsvt} directly process 3D voxels derived from input point clouds.
3D object detectors commonly use Cartesian coordinates as input features.

There exists a handful of studies that, in addition to utilizing Cartesian coordinates, also incorporates local point cloud geometry.
VoxelNet~\cite{zhou2018voxelnet} augments the input data by appending the relative distance between each point and its voxel's centroid to its global coordinates.
Instead of voxel's centroid, PointPillars~\cite{lang2019pointpillars} utilizes distance to the voxel center.
To capture more local information, several methods~\cite{zhou2018voxelnet, wang2023dsvt, lang2019pointpillars} employ a shared PointNet~\cite{Qi_2017_CVPR} architecture to encode pillars or voxels.
\cite{zhao2021integration,miao20213d} utilize a two stream approach, encoding Cartesian coordinates and surface normals in separate branches and fusing them in later stages.
Hybrid methods, which combine local and global geometric inputs, share a common limitation with methods trained solely on global features: they often exhibit poor generalization across diverse domains, likely because their performance is heavily influenced by global features.

In contrast, our proposed approach is \emph{exclusively} based on local point cloud geometry, demonstrating the crucial role of these features in model generalization.
We show that our representation provides more detail (than, \eg, simple relative distance) and is less susceptible to noise (which degrades performance of surface normal-based features).

\nbfnodot{Unsupervised Domain Adaptation (UDA)} transfers knowledge from a labeled source domain to an unlabeled target domain, using both domains to adapt models to the novel data distributions.
Substantial efforts have focused on explicitly addressing different challenges, \eg object size bias~\cite{wang2020trainingermany,3dda:st3d,wang2020trainingermany,malic2023sailor} or overcoming varying LiDAR resolutions~\cite{hu2023density,wei2022lidar,rist2019cross}.
In UDA, self-training remains the dominant paradigm:
This typically involves a closed-loop process where the pseudo-label database is iteratively updated with predictions generated by a model trained on these pseudo-labels~\cite{chen2023revisiting,yang2022st3d++,3dda:st3d,Caine2021,zhang2024pseudo}.
To enhance pseudo-label quality, several methods exploit the temporal dimension inherent in autonomous driving data~\cite{you2022exploiting,FruhwirthReisinger2021,Walsh2020}.

Despite a comprehensive portfolio of LiDAR-based UDA approaches, the impact of using different input features on model performance remains largely unexplored.
Most existing methods rely solely on the standard global point cloud inputs, neglecting other feature channels.
Given the prevalence of self-training in UDA, good domain generalization is crucial to obtain improved pseudo labels.
We demonstrate how our proposed local feature representation significantly enhances the domain generalization of various models, which could be leveraged for improved UDA via common self-training loops.

\nbfnodot{Domain Generalization (DG)}
aims to train models that can effectively generalize to new, unseen domains without requiring any additional data from those domains.
For 2D image-based tasks, DG research primarily focuses on learning domain-invariant features~\cite{rame2022fishr,gong2019dlow,li2018cvprdg,li2018aaaidg} or disentangling domain-specific and domain-invariant features~\cite{lv2022causality,mahajan2021domain,peng2019domain}.
More recently, studies such as \cite{addepalli2024leveraging,chen2024practicaldg} have sought to enhance generalization by employing powerful vision-language models.
A smaller subset of DG research explores data augmentation techniques to improve generalization~\cite{huang2021fsdr,volpi2019addressing,yue2019domain}.

In the context of LiDAR-based detection, data augmentation~\cite{leng2023lidar,hahner2020aug} is crucial to prevent model overfitting to training data.
To this end, \threedvf~\cite{lehner20223d} employs adversarial augmentation to generate diverse and realistic training examples from a single source domain.
In addition to a novel augmentation strategy, Li \etal~\cite{li2023domainne2021} enhance the model robustness by incorporating an auxiliary adversarial task during training and at test-time.
Similarly, MDT3D~\cite{soum2023mdt3d} leverages multiple source domains and cross-dataset augmentation to learn more robust models.
Wu \etal~\cite{wu2023towards} also utilize multiple source domains to learn more general, dataset-invariant features.

A shared characteristic of DG methods for LiDAR-based object detection is their reliance solely on global point coordinates as network inputs.
While the intensity channel has been shown to have a negative impact on domain generalization~\cite{jaeyeul2024eccv,soum2023mdt3d}, the influence of local geometric features has not been explored so far.
Our simple yet effective local feature representation significantly enhances model generalization without requiring tailored data augmentation, auxiliary training tasks, nor test-time adaptation, demonstrating that a focus on local geometric information can be more beneficial than complex DG strategies.

\section{Explicit Local Structure for Better DG}
\label{sec:method}

LiDAR Domain Generalization (DG) is a critical research area for improving the robustness of detection models across domains.
Prior work often relies solely on global point cloud features (\ie Cartesian coordinates) and techniques like data augmentation~\cite{soum2023mdt3d,lehner20223d} or auxiliary tasks~\cite{li2023domainne2021}.
In contrast, we propose leveraging local point cloud geometry to improve the generalization capabilities.

Local features force models to learn object shapes independently of their global position, thereby reducing sensitivity to rigid transformations.
This shape-based representation enhances the model's ability to generalize to new domains.
However, common local features, such as relative distance~\cite{deng2021voxel} or RepSurf~\cite{ran2022surface}, are often used in conjunction with the global representation, hindering their domain generalization effect.
While highly efficient, relative distance features are typically outperformed by more sophisticated representation of local geometry, such as RepSurf (as shown in our experiments).
However, RepSurf leverages surface normal estimates, which are sensitive to noise (due to ambiguity in surface orientation~\cite{hoppe1992surface,li2023neuralgf}), and requires point cloud sorting, causing it to be a computationally expensive input representation.

\begin{figure}
    \centering
    \includegraphics[width=\columnwidth]{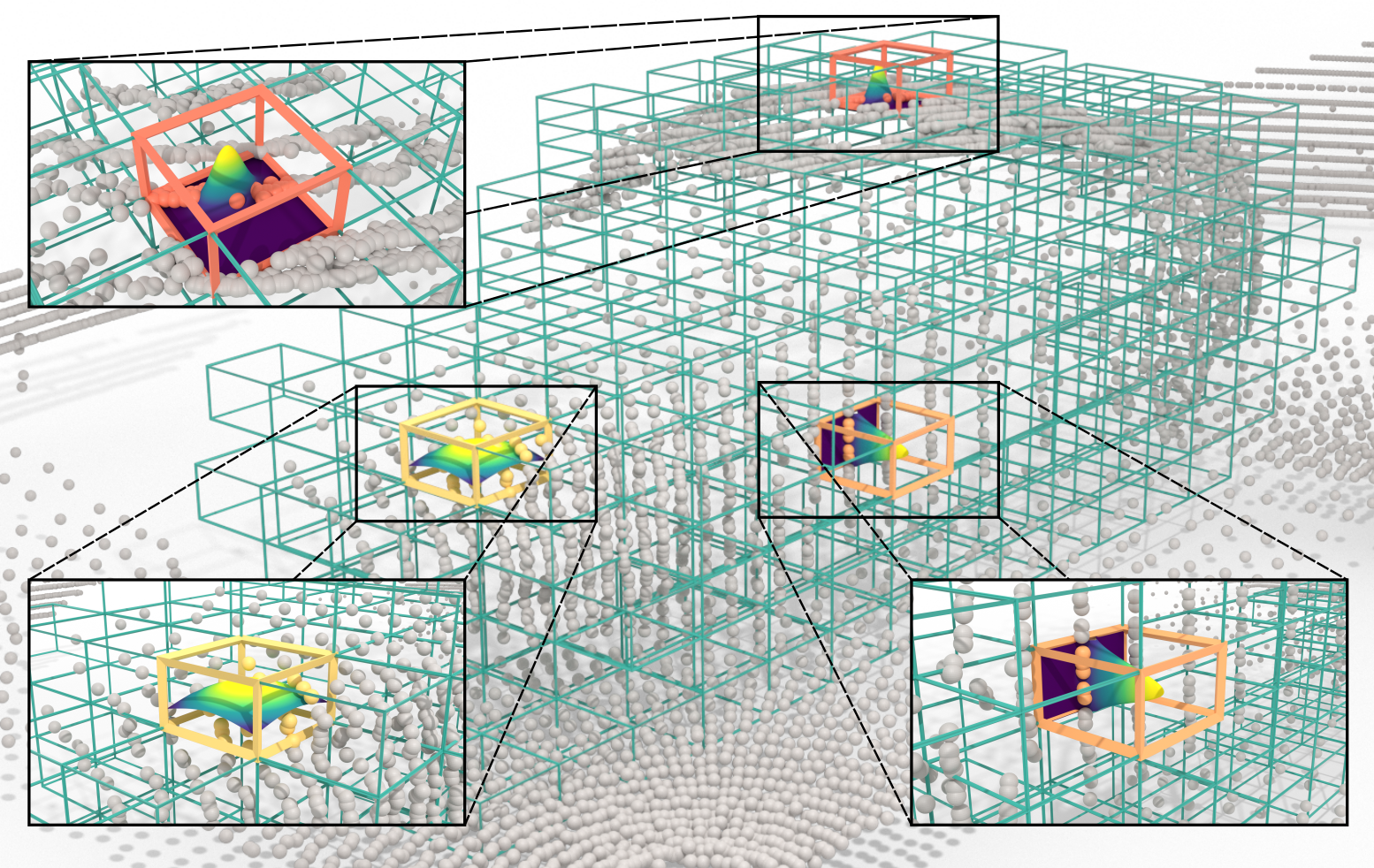}
    \hfil
    \caption{
        Illustration of our \ourmethod encoding:
        We explicitly model the local geometry within point neighborhoods by Gaussian blobs (highlighted).
        Here, shown for a car's point cloud representation, all points within a voxel belong to the same neighborhood.
    }
    \label{fig:method}
\end{figure}

To mitigate these drawbacks, we propose modeling small local neighborhoods of LiDAR point clouds as individual Gaussian blobs, as illustrated in \cref{fig:method}.
Our approach offers several advantages: it preserves local structure more effectively than simple relative distance, thanks to its covariance modeling.
Additionally, it is permutation-invariant, meaning it does not require point cloud sorting, and it avoids the overhead of surface normal calculations, resulting in exceptional efficiency.

\subsection{Preliminaries}
\label{sec:method preliminaries}
Formally, given a set of $D \geq 2$ domains $\{(X_d, Y_d)\}_{d=2}^D$, each with distinct data and label distributions, we train the model on a subset of $K$ domains and conduct the evaluation on a single domain.
We adopt the prevalent assumption that the domain gap primarily stems from discrepancies in data distributions $X_d$, while label distributions $Y_d$ across domains are considered largely consistent.
We then distinguish two different tasks and evaluation protocols: (1) Single Domain Generalization (SDG)~\cite{li2023domainne2021,lehner20223d} when $K=1$, and (2) Multi-Domain Generalization (MDG)~\cite{soum2023mdt3d} when $K=D-1$.

A LiDAR point cloud is represented as an unordered set of points, typically defined as $\{p_j = (x, y, z, i) \}_{j=1}^{M} \sim X_d$, where each point $p_j$ has a 3D coordinate $(x, y, z)$ relative to the sensor's origin and an associated intensity value $(i)$.
Prior research on DG consistently overlooked additional explicit and implicit LiDAR channels, relying solely on $(x, y, z)$ as model input.
For instance, due to the inherent challenges posed by intensity data, it is commonly excluded from the input features~\cite{jaeyeul2024eccv}.
Similarly, the influence of different input descriptors, like relative distances~\cite{deng2021voxel,wang2023dsvt}, surface normals~\cite{zhao2021integration,miao20213d}, or local encoding~\cite{zhou2018voxelnet, wang2023dsvt, lang2019pointpillars}, remains severely underexplored.

\subsection{Local Point Clouds Structure as Gaussian Blobs}
\label{sec:method 3d gaussian rep}

Previous approaches conveniently train a 3D detection model on $K$ domains $\{(X_d, Y_d)\}_{d=2}^K$, using global point positions, $\{p_j = (x, y, z)\}_{j=1}^M \sim X_d$, as input features.
The models are subsequently evaluated on the remaining domain using the same input representation.

Training with such global input features biases the detector towards the source domain(s): Different reference coordinate systems (\eg compare KITTI~\cite{geiger2012kitti} \vs Waymo~\cite{sun2020wod}) lead to different distributions of point coordinates, which are inevitably picked up during training and often hinder the generalization performance.
Thus, it is common practice in both, DG~\cite{soum2023mdt3d,wu2023towards,lehner20223d} and UDA~\cite{chen2023revisiting,yang2022st3d++} approaches, to shift the datasets in the height axis in order to manually align their coordinate distributions.

Conversely, we decouple the absolute object location from model encoding while preserving spatial information by representing a small local point cloud neighborhood as a Gaussian blob, $\mathcal{N} (\bm{\mu}, \Sigma)$, where
\begin{equation}
  \bm{\mu} = \frac{1}{N} \sum_{i=1}^{N} \bm{p}_i, \quad\text{and}
  \label{eq:gauss mean}
\end{equation}
\begin{equation}
  \Sigma = \frac{1}{N} \sum_{i=1}^{N} (\bm{p}_i - \bm{\mu}) (\bm{p}_i - \bm{\mu})^\top \text{.}
  \label{eq:gauss cov}
\end{equation}
The neighborhood of $N$ points is determined by the detector.
For instance, voxel-based detectors naturally limit this neighborhood to the maximum points per voxel.
The dimension of $\bm{\mu} \in \mathbb{R}^M$ and $\Sigma \in \mathbb{R}^{M \times M}$ depends on the dimension of the LiDAR points $p_j \in \mathbb{R}^M$.
Although our representation can leverage intensity information by setting $M=4$ during training, we conduct all our experiments using $M=3$ to ensure a fair comparison with \sota DG methods.
These methods typically exclude intensity information due to its adversarial impact on cross-domain performance, as noted in prior works \cite{jaeyeul2024eccv,soum2023mdt3d}.
In total, the dimensionality of our feature vector is $M+M^2 = 12$.

While $\Sigma$ is decoupled from the absolute point position, $\bm{\mu}$ remains expressed in the LiDAR's coordinate reference frame.
Consequently we simply express $\bm{\mu}$ in a local frame of reference
\begin{equation}
  \bm{d} = \frac{1}{N} \sum_{i=1}^{N} \bm{p}_i - \bm{\mu} \text{.}
  \label{eq:loc d}
\end{equation}
We directly use $(\bm{d}, \Sigma)$ as an input to an arbitrary model.
\cref{alg:3d gaussian blob} summarizes our Gaussian blob computation in pseudocode.
\begin{algorithm}[t]
\caption{\ourmethod in a PyTorch-like pseudocode.}
\label{alg:3d gaussian blob}
\definecolor{codeblue}{rgb}{0.25,0.5,0.5}
\lstset{
  backgroundcolor=\color{white},
  basicstyle=\fontsize{7.2pt}{7.2pt}\ttfamily\selectfont,
  columns=fullflexible,
  breaklines=true,
  captionpos=b,
  commentstyle=\fontsize{7.2pt}{7.2pt}\color{codeblue},
  keywordstyle=\fontsize{7.2pt}{7.2pt},
}
\begin{lstlisting}[language=python,escapechar=\%]
# f: N queries, each containing K neighborhood points of dimension M (NxKxM)

# compute mean %\hfill$\smalltriangleright$ \cref{eq:gauss mean}%
f_mean = f.mean(dim=1, keepdims=True)
# compute cov %\hfill$\smalltriangleright$ \cref{eq:gauss cov}%
f_loc = f - f_mean
cov = torch.einsum("nka,nkb->nab", f_loc, f_loc)
# final descriptor
d = f_loc.mean(dim=1) # NxM %\hfill$\smalltriangleright$ \cref{eq:loc d}%
cov_flat = cov.reshape(-1, pow(M, 2)) # NxM^2
gblobs = torch.cat([d, cov_flat], dim=1) # (Nx(M+M^2))
\end{lstlisting}
\end{algorithm}

\subsection{Global \vs Local Representations}
\label{sec:method discussion}
The common practice of using global point cloud coordinates as input features in 3D models is analogous to appending pixel coordinates to RGB values of images.
This, however, would violate the assumed translation invariance, because the same object at different locations would have distinct feature activations.
Moreover, the explicit global representation is unnecessary, as positional information is already implicitly encoded in 2D/3D CNNs~\cite{liu2022convnet,yan2018second}.
Even transformer architectures~\cite{alexey2020image,vaswani2017attention,liu2021swin,wang2023dsvt}, which lack an inherent notion of position, explicitly inject it through different positional encodings.
Therefore, the use of absolute position as additional input features is redundant and, as we show, can be detrimental for generalization performance.

Our local geometry $(\bm{d}, \Sigma)$ effectively separates an object's spatial position within a LiDAR frame from its encoding.
Unlike global coordinates, similar local geometry induces similar activations, regardless of their absolute location.
By explicitly representing the local geometry, we leave the positional information to the network architecture.
This disentanglement allows the model to learn more abstract and invariant representations of objects, effectively replacing the redundant position information with a geometrically sound representation.

\section{Experiments}
\label{sec:experiments}
We present detailed experiments to validate our approach on widely-used benchmarks.
\cref{sec:experiments single source} and \cref{sec:experiments multi source} present our findings from single-source~\cite{li2023domainne2021} and multi-source~\cite{soum2023mdt3d} domain generalization experiments, respectively.
Finally, \cref{sec:ablation studies} presents our thorough ablation studies.

\nbf{Datasets}
In order to be comparable with the existing benchmarks~\cite{li2023domainne2021,soum2023mdt3d,lehner20223d}, we employ four commonly used autonomous driving datasets, as outlined in \cref{tab:datasets}.
The variety of datasets allows us to thoroughly validate our claims.
Differences in the number of LiDAR beams and vertical field-of-view induce vastly different numbers of points per-sample (\ie per LiDAR point cloud).
Diverse LiDAR configurations create unique sampling patterns, biasing detection models towards their source data.
This is evident in their poor cross-domain performance, as demonstrated in our subsequent evaluations.
\begin{table}
\footnotesize
\centering
\resizebox{\columnwidth}{!}{%
\begin{tabular}{llccc}
\toprule
\textbf{Dataset} & \textbf{Location} & \textbf{\# Beams} & \textbf{PPS}   & \textbf{VFOV}                          \\
\midrule
KITTI~\cite{geiger2012kitti}            & Germany           & 64                & 118k           & $ [-23.6^\circ, \phantom{0}3.2^\circ]$ \\
Waymo~\cite{sun2020wod}            & USA               & 64                & 160k           & $[-17.6^\circ, \phantom{0}2.4^\circ]$  \\
nuScenes~\cite{caesar2020nuscenes}         & USA/Singapore     & 32                & \phantom{0}25k & $[-30.0^\circ, 10.0^\circ]$            \\
ONCE~\cite{mao2022one}             & China             & 40                & \phantom{0}70k & $[-25.0^\circ, 15.0^\circ]$            \\
\bottomrule
\end{tabular}}
\caption{
  Characteristics of datasets used in our experiments, including their geographic location, number of LiDAR beams, points per-scan (PPS), and vertical field-of-view.
}
\label{tab:datasets}
\end{table}

\nbf{Metrics}
Following the established evaluation protocols~\cite{yang2022st3d++,3dda:st3d}, we use the KITTI~\cite{geiger2012kitti} evaluation metrics in all our experiments.
If not stated otherwise, we utilize three main classes: Car (\ie Vehicle for Waymo), Pedestrian and Cyclist.
We report Average Precision (AP) computed over 40 recall points and mean AP (mAP) over all classes.
The Intersection over Union (IoU) thresholds are set at $0.7$, $0.5$ and $0.5$ for Car, Pedestrian and Cyclists, respectively.

\nbf{Baselines}
As a baseline, we report the performance of a model trained with default configuration, \ie, its standard training settings (learning rate, number of epochs, \etc) and standard 3D data augmentations (ground truth sampling, rotation, scaling, and flipping).
We compare this model with \sota single-source~\cite{choi2021part,lehner20223d,li2023domainne2021} and multi-source~\cite{soum2023mdt3d} domain generalization (DG) methods.
To demonstrate our contributions, we train the default model again but replace the global inputs with our proposed method, as described in \cref{sec:method}.
For this, we do not employ any specialized data augmentation techniques, model parameter search, or hyperparameter optimization.
For the evaluation, we simply select the last checkpoint.

\subsection{Single-source Domain Generalization}
\label{sec:experiments single source}

Single-source domain generalization aims to train a model on a single source domain and generalize its performance to unseen target domains.
In our experiments, we follow established benchmarks~\cite{yang2022st3d++,3dda:st3d,li2023domainne2021,lehner20223d} and test our method on three different dataset configurations: \WtoK, \NtoK and \WtoN.
For apples-to-apples comparisons with other DG methods~\cite{choi2021part,lehner20223d,li2023domainne2021,li2023domainne2021}, we use the same detector and point cloud configuration:
We train Voxel R-CNN~\cite{deng2021voxel}, limit the LiDAR range to $[-75.2m, -75.2m, -2m, 75.2m, 75.2m, 4m]$ and utilize a voxel size of $[0.1m, 0.1m, 0.15m]$.
To train the model, we leveraged the OpenPCDet\footnote{\url{https://github.com/open-mmlab/OpenPCDet}} framework.

We compare our method to PA-DA~\cite{choi2021part} and \threedvf~\cite{lehner20223d}, which use different specialized data augmentation techniques to improve model generalization.
Additionally, we also benchmark our method against the \sota by Li~\etal~\cite{li2023domainne2021}, which uses a multi-task learning strategy in addition to a novel data augmentation technique.
Moreover, they exploit a point cloud reconstruction task for test-time training.
On the contrary, our method does not rely on heavy data augmentation, additional training tasks, or test-time training.
We simply train the baseline model with our \ourmethod input features, using the default configuration.
Our simple, yet highly effective approach surpasses existing DG approaches as shown in \cref{tab:single source domain generalization}, achieving substantial improvements of over $21$ and $12$ mAP in \WtoK and \NtoK, respectively.
While the relatively small improvement observed in the \WtoN experiment is discussed in detail in our ablation studies, the detector's inability to bridge the dense-to-sparse gap -- a well-known challenge in 3D object detection~\cite{wei2022lidar,eskandar2024empirical} -- remains a primary limitation.
\begin{table*}[t]
\centering
\begin{tabular}{llcccc}
\toprule
\textbf{Tasks}                                & \textbf{Methods}                   & \textbf{Car}                                             & \textbf{Pedestrian}                                      & \textbf{Cyclist}                                                             & \textbf{mAP}                                             \\
\midrule
\multirow{5}{*}{Waymo $\rightarrow$ KITTI}    & Voxel R-CNN~\cite{deng2021voxel}                        & 66.65/19.27                                              & 66.55/64.00                                              & 63.04/57.11                                                                  & 65.41/46.79                                              \\
                                              &PA-DA\cite{choi2021part}           & 65.82/17.61                                              & 66.40/63.88                                              & 61.30/56.23                                                                  & 64.51/45.91                                              \\
                                              &3D-VF\cite{lehner20223d}           & 66.72/19.37                                              & 66.21/63.12                                              & 62.74/56.44                                                                  & 65.22/46.31                                              \\
                                              &Li~\etal~\cite{li2023domainne2021} & 69.90/20.21                                              & 63.24/62.59                                              & 63.27/57.21                                                                  & 65.47/46.67                                              \\
                                              & \cellcolor{gray!25}Voxel R-CNN w/ \ourmethod            & \cellcolor{gray!25}\bestresult{87.33}/\bestresult{78.75} & \cellcolor{gray!25}\bestresult{69.47}/\bestresult{65.98} & \cellcolor{gray!25}\bestresult{63.63}/\bestresult{58.72}                     & \cellcolor{gray!25}\bestresult{73.48}/\bestresult{67.82} \\
\midrule

\multirow{5}{*}{nuScenes $\rightarrow$ KITTI} & Voxel R-CNN~\cite{deng2021voxel}                        & 66.93/28.80                                              & 23.39/18.65                                              & 19.23/15.76                                                                  & 36.52/21.07                                              \\
                                              & PA-DA~\cite{choi2021part}          & 65.09/32.44                                              & 18.73/14.94                                              & 18.66/15.91                                                                  & 34.16/21.10                                              \\
                                              & 3D-VF~\cite{lehner20223d}          & 65.36/29.21                                              & 24.85/20.87                                              & 22.13/19.31                                                                  & 37.45/23.13                                              \\
                                              & Li~\etal~\cite{li2023domainne2021} & 73.58/33.11                                              & 30.01/23.73                                              & 22.93/18.62                                                                  & 42.17/25.15                                              \\
                                              & \cellcolor{gray!25}Voxel R-CNN w/ \ourmethod            & \cellcolor{gray!25}\bestresult{80.95}/\bestresult{53.98} & \cellcolor{gray!25}\bestresult{38.33}/\bestresult{33.22} & \cellcolor{gray!25}\bestresult{29.18}/\bestresult{25.68}                     & \cellcolor{gray!25}\bestresult{49.48}/\bestresult{37.62} \\
\midrule
\multirow{5}{*}{Waymo $\rightarrow$ nuScenes} & Voxel R-CNN~\cite{deng2021voxel}                        & 31.20/19.13                                              & 10.52/\phantom{0}8.39                                    & \phantom{0}0.75/\phantom{0}0.55                                              & 14.16/\phantom{0}9.36                                    \\
                                              & PA-DA\cite{choi2021part}           & 29.43/18.06                                              & 10.84/\phantom{0}8.43                                    & \phantom{0}0.82/\phantom{0}0.43                                              & 13.70/\phantom{0}8.97                                    \\
                                              & 3D-VF\cite{lehner20223d}           & 30.17/18.91                                              & 10.54/\phantom{0}7.23                                    & \phantom{0}0.76/\phantom{0}0.78                                              & 13.82/\phantom{0}8.97                                    \\
                                              & Li~\etal~\cite{li2023domainne2021} & \bestresult{36.04}/\bestresult{22.25}                    & \bestresult{14.48}/\bestresult{10.56}                    & \phantom{0}1.15/\phantom{0}0.95                                              & \bestresult{17.22}/11.26                                 \\
                                              & \cellcolor{gray!25}Voxel R-CNN w/ \ourmethod            & \cellcolor{gray!25}32.08/20.08                           & \cellcolor{gray!25}11.60/\phantom{0}9.13                 & \cellcolor{gray!25}\bestresult{\phantom{0}5.67}/\bestresult{\phantom{0}5.10} & \cellcolor{gray!25}16.45/\bestresult{11.44}              \\
\bottomrule
\end{tabular}
\caption{
  Single-source domain generalization experiments using Voxel R-CNN~\cite{deng2021voxel}.
  Following Li \etal~\cite{li2023domainne2021}, we trained a Voxel R-CNN detector on all three classes (Car/Vehicle, Pedestrian, Cyclist) simultaneously and evaluated performance using Average Precision (AP) on Bird's-eye View (BEV) / 3D views at $40$ recall positions.
  Intersection over Union (IoU) thresholds of $0.7$, $0.5$, and $0.5$ were used for Car/Vehicle, Pedestrian, and Cyclist, respectively.
  For KITTI evaluation, we report the average AP across all difficulty levels (Easy, Moderate, Hard).
  Additionally, we provide the mean AP over the three classes.
  The best value in each category is highlighted in bold.
}
\label{tab:single source domain generalization}
\end{table*}

For additional insights, we replicate the \KtoW evaluation from 3D-VF~\cite{lehner20223d}:
We train three detectors (PointPillars~\cite{lang2019pointpillars}, SECOND~\cite{yan2018second}, and Part-A$^2$~\cite{shi2020points}) using their default configurations (see supplementary material for details) and simply exchange their default global input features with our \ourmethod.
We evaluate these models on the full KITTI (in-domain) and Waymo eval split, using the Car/Vehicle class. 
Analogous to \threedvf, we report AP with a 3D IoU of $0.7$ for KITTI and $0.5$ for Waymo.
As shown in \cref{tab:single source domain generalization 3d-vfield}, our input encoding outperforms 3D-VF by over $4$, $13$, and $4$ AP points using PointPillars, SECOND, and Part-A$^2$, respectively.
Our proposed input encoding and standard detector settings were sufficient to achieve significant improvement without any specialized data augmentation.
We observed that data augmentation tailored for the target domain can often negatively impact performance on the source domain, as demonstrated in the KITTI evaluation of \cref{tab:single source domain generalization 3d-vfield}.
In contrast, our approach is unaffected by this issue and even outperforms other methods in in-domain evaluation.
\begin{table}[t]
\small
\centering
\begin{tabular}{llcc}
\toprule
\textbf{Model}                                         & \textbf{Method}           & \textbf{KITTI}                        & \textbf{Waymo}                        \\
\midrule
\multirow{3}{*}{PointPil.~\cite{lang2019pointpillars}} & Default               & 77.11                                 & 40.86                                 \\
                                                       & 3D-VF~\cite{lehner20223d} & 77.13                                 & 44.61                                 \\
                                                       & \cellcolor{gray!25}w/ \ourmethod   & \cellcolor{gray!25}\bestresult{78.75} & \cellcolor{gray!25}\bestresult{48.92} \\
\midrule
\multirow{3}{*}{SECOND~\cite{yan2018second}}           & Default               & 78.68                                 & 42.45                                 \\
                                                       & 3D-VF~\cite{lehner20223d} & 78.56                                 & 43.51                                 \\
                                                       & \cellcolor{gray!25}w/ \ourmethod   & \cellcolor{gray!25}\bestresult{81.61} & \cellcolor{gray!25}\bestresult{56.52} \\
\midrule
\multirow{3}{*}{Part-A$^2$~\cite{shi2020points}}       & Default               & 79.16                                 & 49.76                                 \\
                                                       & 3D-VF~\cite{lehner20223d} & 79.26                                 & 56.08                                 \\
                                                       & \cellcolor{gray!25}w/ \ourmethod   & \cellcolor{gray!25}\bestresult{82.29} & \cellcolor{gray!25}\bestresult{60.20} \\

\bottomrule
\end{tabular}
\caption{
  Following \threedvf~\cite{lehner20223d}, we trained three detectors on the KITTI dataset (Car class only), and evaluated them on KITTI (Moderate difficulty, IoU threshold 0.7) and Waymo (IoU threshold 0.5) using 3D Average Precision (AP).
}
\label{tab:single source domain generalization 3d-vfield}
\end{table}

\subsection{Multi-source Domain Generalization}
\label{sec:experiments multi source}

Multi-source DG trains models that can effectively generalize to unseen data from different distributions.
By leveraging multiple source domains, the model can capture underlying patterns and invariant features that are common to all domains, enabling it to perform well on novel, unseen target domains.
This approach is particularly valuable in scenarios where target domain data is limited or unavailable, making it a crucial tool for real-world applications, especially in fields like autonomous driving.

Given the limited research on LiDAR-based 3D object detection in multi-source domain generalization, we benchmark our approach against the \sota MDT3D~\cite{soum2023mdt3d}.
Following their leave-one-out strategy (where one dataset is used for testing and the others for training), we train a CenterPoint~\cite{yin2021center} model on three of the four widely used autonomous driving datasets (KITTI~\cite{geiger2012kitti}, Waymo~\cite{sun2020wod}, nuScenes~\cite{caesar2020nuscenes}, and ONCE~\cite{caesar2020nuscenes}) and evaluate on the remaining one.
To ensure a fair comparison with MDT3D, we apply their uniform data subsampling schema and use the same LiDAR range $[-75.2m, -75.2m, -2m, 75.2m, 75.2m, 4m]$ and voxel size $[0.1m, 0.1m, 0.2m]$.
Detailed experiment configurations are provided in the supplemental material.

In \cref{tab:multi source domain generalization mdt3d centerpoint}, we report the mean Average Precision (mAP) computed over all classes for each dataset.
Following~\cite{soum2023mdt3d}, for KITTI, we report the moderate difficulty, whereas for the others there is no such categorization.
Unlike MDT3D, which uses multi-domain data mixing, our method employs the standard CenterPoint configuration with default augmentations (ground truth sampling, rotation, scaling, and flipping) and no additional hyperparameter tuning.
Without any bells and whistles our method exhibits tremendous improvements over MDT3D of around $17$, $12$ and $5$ mAP for Waymo, KITTI and ONCE.
Similar to \cref{sec:experiments single source}, extremely sparse neighborhoods (\ie nuScenes) poses a challenge, as we further investigate within our ablation studies.
\begin{table*}
\centering
\begin{tabular}{lccccc}
\toprule
\textbf{Method}            & \textbf{KITTI}                        & \textbf{ONCE}                                          & \textbf{nuScenes}       & \textbf{Waymo} & \textbf{mAP}                        \\
\midrule
KITTI                      & \textcolor{lightgray}{47.40}                                 & \phantom{0}9.90\phantom{$^\dag$}                       & \phantom{0}1.60         & \phantom{0}3.00 & \phantom{0}4.83                       \\
ONCE                       & 43.00                                 & \textcolor{lightgray}{60.20}\phantom{$^\dag$}                                 & \phantom{0}5.80         & 13.00 & 20.60                                \\
nuScenes                   & 12.60                                 & 10.70\phantom{$^\dag$}                                 & \textcolor{lightgray}{18.40}                   & \phantom{0}0.90 & \phantom{0}8.06                      \\
Waymo                      & 34.20                                 & 28.80\phantom{$^\dag$}                                 & \phantom{0}7.20         & \textcolor{lightgray}{40.20} & 23.40\\
\midrule
MDT3D~\cite{soum2023mdt3d} & 41.80                                 & 32.28$^\dag$                                           & \bestresult{11.00}      & \phantom{0}6.40 & 22.87                      \\
\cellcolor{gray!25}CP w/ \ourmethod    & \cellcolor{gray!25}\bestresult{53.92} & \cellcolor{gray!25}\bestresult{37.77}\phantom{$^\dag$} & \cellcolor{gray!25}\phantom{0}8.15 & \cellcolor{gray!25}\bestresult{23.81} & \cellcolor{gray!25}\bestresult{30.91} \\
\bottomrule
\end{tabular}
\caption{
    Multi-source DG results.
    The top part presents the results for a vanilla CenterPoint~\cite{yin2021center} (CP) detector, trained exclusively on single domains (leftmost column) and evaluated on all datasets.  
    For each dataset, we report mean Average Precision (mAP) over all classes (columns KITTI--Waymo) and across all datasets, excluding in-domain experiments (rightmost column).
    The bottom part presents the multi-source DG results following MDT3D's leave-one-out strategy, \ie CenterPoint is trained with our \ourmethod as input features on three datasets and evaluated on the fourth.
    $^\dag$:  in Tab. 7 of MDT3D~\cite{soum2023mdt3d}, the mAP is wrongly calculated; we report the corrected value.
}
\label{tab:multi source domain generalization mdt3d centerpoint}
\end{table*}

\subsection{Ablation Studies}
\label{sec:ablation studies}

\nbf{Input Feature Impact}
In \cref{tab:component_influence}, we compare the performance of models trained on different input features for in- and cross-domain evaluation.
Our results show that the choice of input features has little impact on the in-domain performance.
However, local features significantly outperform global and hybrid (combined local and global) features in the cross-domain scenario.
There, our proposed \ourmethod, \ie $(\mathbf{b}, \Sigma)$, achieve the best result.
\begin{table}
\footnotesize
\centering
\begin{tabular}{ccccc}
\toprule
global                             & $\mathbf{d}$  & $\Sigma$ & n$\rightarrow$n & n$\rightarrow$K \\
\midrule
\checkmark & & & \bestresult{30.21}/\bestresult{22.49} & 36.52/21.07 \\
 \checkmark & & \checkmark & 29.79/22.37 & 36.58/22.45 \\
 & \checkmark & & 29.29/21.84 & 48.61/35.95 \\
& & \checkmark & 28.51/21.13 & 49.02/37.60 \\
& \checkmark & \checkmark & 29.33/21.90 & \bestresult{49.48}/\bestresult{37.62} \\
\bottomrule
\end{tabular}
\caption{
    Impact of different input features on Voxel R-CNN~\cite{deng2021voxel}.
    We evaluate the in-domain nuScenes$\rightarrow$nuScenes (n$\rightarrow$n) and cross-domain \NtoK (n$\rightarrow$K) performance.
    We report mAP, computed over all three classes, on BEV / 3D views at $40$ recall positions.
    We denote standard training with global features (\ie Cartesian coordinates) as "global".
    $\mathbf{d}$ and $\Sigma$ are features computed as in \cref{eq:loc d} and \cref{eq:gauss cov}, respectively.
}
\label{tab:component_influence}
\end{table}

\nbf{In-domain Performance}
While \threedvf~\cite{lehner20223d} found varying levels of in-domain performance among different 3D detectors (similar to our \cref{tab:single source domain generalization 3d-vfield}), Li \etal~\cite{li2023domainne2021} and MDT3D~\cite{soum2023mdt3d} do not report any in-domain evaluation results.
Given that their data augmentation strategies are focused on target datasets, it is likely that their model performance on the source dataset would degrade, although the extent is uncertain.
While our Gaussian blobs outperform these methods in competitive domain generalization (DG) benchmarks, the impact of our approach on in-domain performance remains unclear.
To assess this, we evaluated various detectors on different autonomous driving datasets, as shown in Table \ref{tab:in-domain}.
Our experiments on the KITTI dataset (\cref{tab:in-domain kitti}) show that our input features surpass the default detector in all classes.
Additionally, our method improves a DSVT (Pillar) transformer~\cite{wang2023dsvt} by over $3$ mAP on the Waymo dataset (\cref{tab:in-domain waymo}).
\begin{table}
\begin{subtable}{\columnwidth}
\footnotesize
\centering
\resizebox{\columnwidth}{!}{%
\begin{tabular}{lcccc}
\toprule
\textbf{Method}                 & \textbf{Car}                        & \textbf{Pedestrian}                 & \textbf{Cyclist}                    & \textbf{mAP}                        \\
\midrule
SEC.~\cite{yan2018second}     & 88.2/80.9                           & 54.2/49.8                           & 68.0/63.1                           & 70.1/64.6                           \\
SEC. w/ \ourmethod                            & \bestresult{88.8}/\bestresult{81.0} & \bestresult{55.7}/\bestresult{50.4} & \bestresult{69.0}/\bestresult{64.5} & \bestresult{71.2}/\bestresult{65.3} \\
\bottomrule
\end{tabular}
}%
\caption{
  In-domain evaluation of SECOND~\cite{yan2018second} on the KITTI \textit{val} split (Moderate case), reporting Average Precision (AP) BEV/3D for IoU thresholds of $0.7$ (Car) and $0.5$ (Pedestrian and Cyclist).
}
\label{tab:in-domain kitti}
\end{subtable}
\begin{subtable}{\columnwidth}
\footnotesize
\centering
\resizebox{\columnwidth}{!}{%
\begin{tabular}{lcccc}
\toprule
\textbf{Method}                 & \textbf{Vehicle}                    & \textbf{Pedestrian}                 & \textbf{Cyclist}                    & \textbf{mAP}                        \\
\midrule
DSVT~\cite{wang2023dsvt} & 74.1/65.7                           & 73.3/65.4                           & 58.0/55.8                           & 68.5/62.3                           \\
DSVT w/ \ourmethod                            & \bestresult{75.4}/\bestresult{67.0} & \bestresult{76.7}/\bestresult{68.8} & \bestresult{64.4}/\bestresult{62.0} & \bestresult{72.2}/\bestresult{65.9} \\
\bottomrule
\end{tabular}
}%
\caption{
  In-domain evaluation on the full Waymo \textit{val} split reporting L1/L2 Average Precision (AP).
  Models are trained on $20\%$ of the training split.
}
\label{tab:in-domain waymo}
\end{subtable}
\caption{
  In-domain experiments on KITTI~\cite{geiger2012kitti} and Waymo~\cite{sun2020wod}.
}
\label{tab:in-domain}
\end{table}

\nbf{Different Local Descriptors}
A limited subset of methods augment point clouds with local geometric features, such as relative distance from voxels~\cite{deng2021voxel} or umbrella-like handcrafted descriptors~\cite{ran2022surface}, in addition to standard global inputs.
However, models trained on such hybrid features inherit all the disadvantages of global geometric features (see \cref{tab:component_influence}).
Therefore, in \cref{tab:local descriptors}, we remove all global components (if there are any) and compare various local point cloud features in the \WtoK case.

Encoding local geometry simply as relative distances is invariant to global transformations and provides good results.
However, this encoding is limited in its ability to capture fine-grained details, which handcrafted descriptors (\eg RepSurf~\cite{ran2022surface}) can provide.
Unlike RepSurf, our method does not require surface normal estimates, which are highly sensitive to noise~\cite{scheuble2024polarization}.
Moreover, it is inherently permutational invariant and does not require any point cloud sorting, yet it consistently outperforms all other methods on the \WtoK DG benchmark.

\begin{table}
\scriptsize %
\centering
\resizebox{\columnwidth}{!}{%
\begin{tabular}{lcccc}
\toprule
\textbf{Method}                             & \textbf{Car}  & \textbf{Pedestrian} & \textbf{Cyclist} & \textbf{mAP} \\
\midrule
rel. distance~\cite{deng2021voxel}              & 84.93/74.60 & 68.72/65.46 & 60.47/55.96 & 71.38/65.34 \\
surf. normals~\cite{hoppe1992surface}         & 85.39/73.37 & 66.32/63.96 & \bestresult{63.96}/58.44 & 70.05/63.96          \\
RepSurf~\cite{ran2022surface}               & 84.45/73.47 & 69.41/\bestresult{66.87} & 61.98/57.70 & 71.95/66.02            \\
\cellcolor{gray!25}\ourmethod            & \cellcolor{gray!25}\bestresult{87.33}/\bestresult{78.75} & \cellcolor{gray!25}\bestresult{69.47}/65.98 & \cellcolor{gray!25}63.63/\bestresult{58.72}                     & \cellcolor{gray!25}\bestresult{73.48}/\bestresult{67.82} \\
\bottomrule
\end{tabular}
}%
\caption{
  \WtoK evaluation of different local descriptors using Voxel R-CNN~\cite{deng2021voxel}.
  We report AP on BEV / 3D views at $40$ recall positions and mAP computed over all three classes.
}
\label{tab:local descriptors}
\end{table}

\begin{figure}
     \centering
     \begin{subfigure}{0.23\textwidth}
         \centering
         \includegraphics[width=\textwidth]{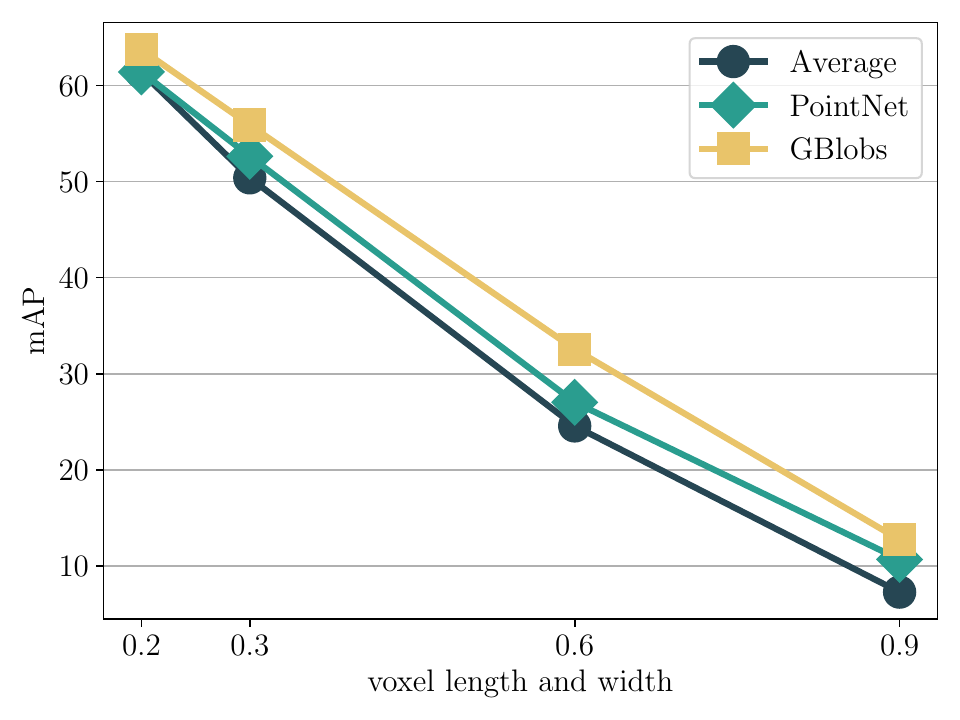}
         \caption{Waymo$\rightarrow$Waymo.}
         \label{fig:information retention w->w}
     \end{subfigure}
     \hfill
     \begin{subfigure}{0.23\textwidth}
         \centering
         \includegraphics[width=\textwidth]{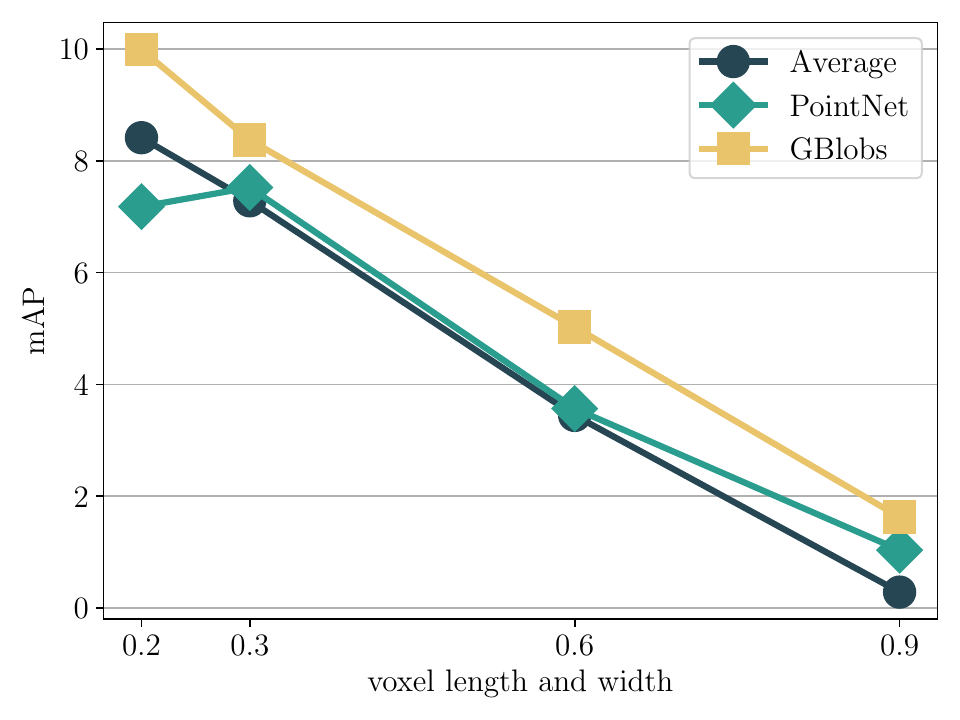}
         \caption{\WtoN.}
         \label{fig:information retention w->n}
     \end{subfigure}
        \caption{
            Impact of voxel size and voxel feature encoder for Voxel R-CNN~\cite{deng2021voxel}, trained on Waymo and evaluated for the (a)~in-domain and (b) cross-domain setting.
            Best viewed on screen.
        }
        \label{fig:information retention}
\end{figure}

\nbf{Information Retention}
Voxel-based detectors (\eg \cite{deng2021voxel,yan2018second}) average points within voxels to create per-voxel features, which can lead to data loss.
Using smaller voxels mitigates this by increasing granularity but at a higher computational cost.
Pillar-based detectors (\eg \cite{li2023pillarnext,lang2019pointpillars}) use voxels spanning the full height, often employing a small PointNet~\cite{Qi_2017_CVPR} to encode pillar features, though some information loss is inevitable.
Our Gaussian blobs outperform existing encoders in preserving information: as shown in \cref{fig:information retention}, larger voxels degrade performance for both in-domain (\cref{fig:information retention w->w}) and cross-domain (\cref{fig:information retention w->n}) evaluations, regardless of the encoder.
Our encoding consistently surpasses PointNet voxel encoders, even with fewer parameters, likely because voxel encoders struggle to reliably encode local structure from global coordinates without additional supervision.

\nbf{nuScenes Performance}
Both single-source (\cref{tab:single source domain generalization}) and multi-source (\cref{tab:multi source domain generalization mdt3d centerpoint}) DG approaches struggle with the challenging nuScenes benchmark.
This dense-to-sparse gap is a well-known challenge in 3D object detection~\cite{wei2022lidar,eskandar2024empirical}.
While our \ourmethod consistently yields significant improvements across all other benchmarks, the achievable performance on nuScenes is on-par with default global input features.

Compared to Waymo and KITTI, nuScenes has significantly fewer points per LiDAR frame (recall \cref{tab:datasets}) and a distribution that is skewed towards the lower range: only about $1.5$k voxels ($6\%$ of the total) contain more than three points, while $94\%$ of all nuScenes voxels contain only one or two points.
The underdetermined system, lacking sufficient data to estimate the covariance matrix as per \cref{eq:gauss cov}, results in a degenerate mean-only case.
This is because the covariance matrix values are forced to zero, significantly limiting the model's ability to capture complex data distributions and leading to poor nuScenes performance.
However, we anticipate that with the constant sensor improvements, such sparse LiDAR point-clouds (\eg sparser than nuScenes) will soon become obsolete, making this generalization scenario highly unlikely in real-world autonomous driving applications. 

\nbf{Sparsity Influence} %
To evaluate the impact of sparsity on our input representation, we conduct two experiments: an in-domain evaluation on KITTI~\cite{geiger2012kitti} using SECOND~\cite{yan2018second} and a cross-domain \NtoK evaluation using Voxel R-CNN~\cite{deng2021voxel}.
The in-domain model was trained using the standard KITTI configuration, while the cross-domain experiment followed the setup described in \cref{sec:experiments single source}.
In both cases, we report mean Average Precision (mAP) computed over the Car, Pedestrian, and Cyclist classes.

\begin{figure}
     \centering
     \begin{subfigure}{0.23\textwidth}
         \centering
         \includegraphics[width=\textwidth]{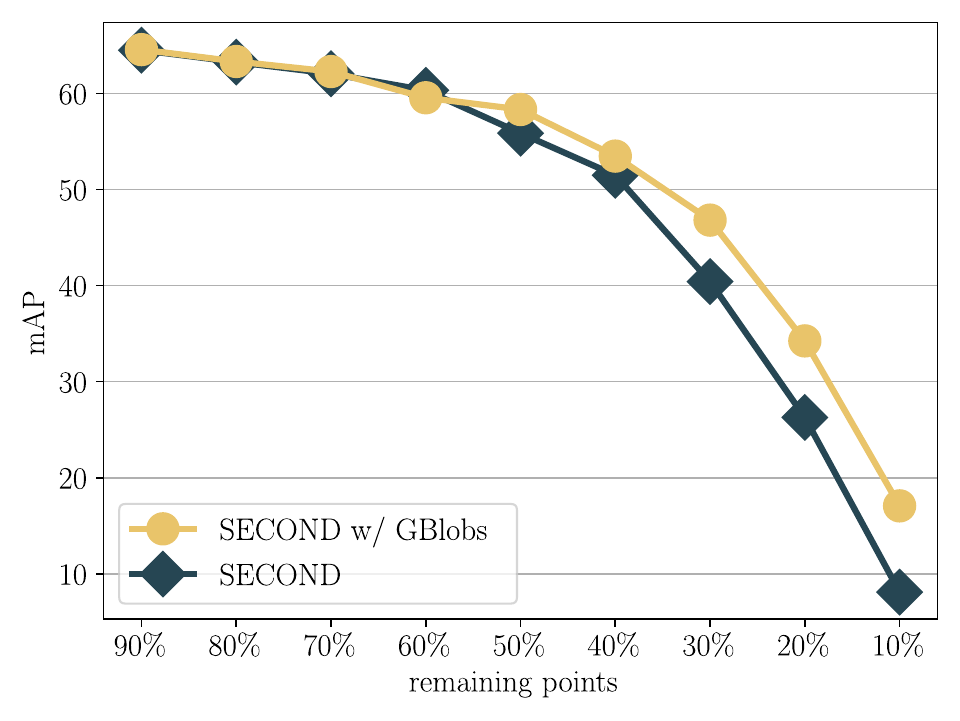}
         \caption{
            In-domain: SECOND~\cite{yan2018second} on KITTI$\rightarrow$"sparser KITTI".
        }
         \label{fig:points subsampling kitti}
     \end{subfigure}
     \hfill
     \begin{subfigure}{0.23\textwidth}
         \centering
         \includegraphics[width=\textwidth]{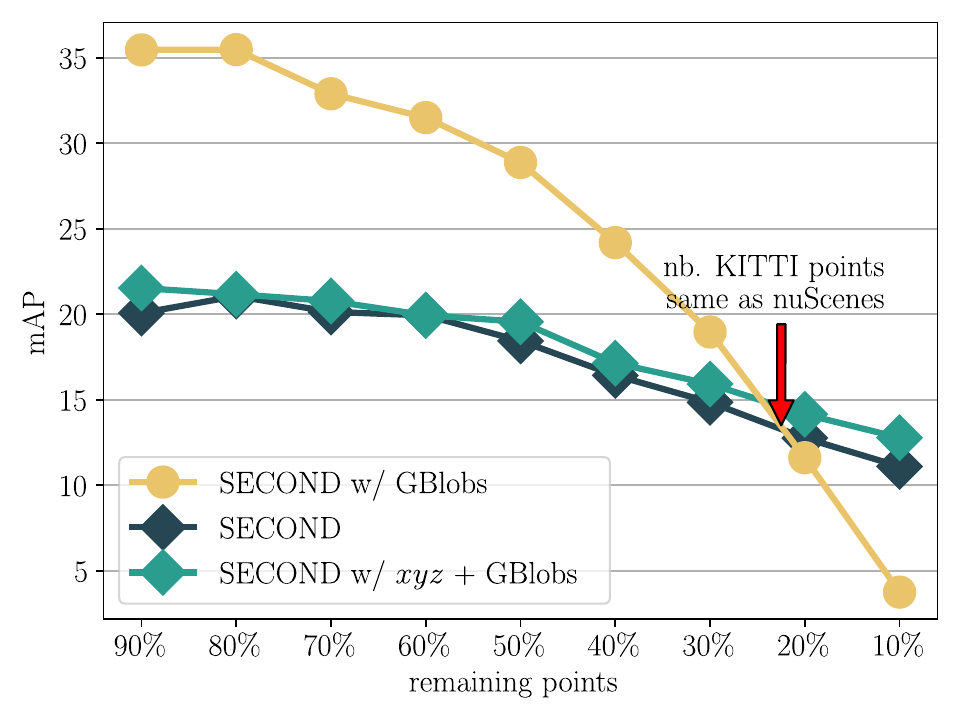}
         \caption{
           Cross-domain: Voxel R-CNN \cite{deng2021voxel} on nuScenes$\rightarrow$"sparser KITTI".
        }
         \label{fig:points subsampling nuscenes to kitti}
     \end{subfigure}
        \caption{
        Extreme sparsity evaluation: detectors trained on complete point clouds were evaluated on gradually sub-sampled KITTI point clouds at test time, both for the (a) in-domain and (b) cross-domain setting. Best viewed on screen.
        }
        \label{fig:points subsampling}
\end{figure}

During testing, we uniformly sampled the input point cloud to retain a specified portion of the points, discarding the rest.
As illustrated in \cref{fig:points subsampling}, a higher point cloud density correlates with improved precision for both in-domain and cross-domain experiments.
Conversely, reducing the number of input points resulted in a decrease in performance.

In \cref{fig:points subsampling kitti}, reducing the number of input points transforms the task into a domain generalization problem.
In such cases, as previously shown, our method excels.
Even when $90\%$ of the data points were discarded, the model trained with our inputs outperformed the original by over $10$ mAP.

Our sparse cross-domain experiment in \cref{fig:points subsampling nuscenes to kitti} reveals that in the extreme \enquote{sparse-to-sparser} case, where the target dataset has far fewer points than the already sparse source dataset, global point cloud features outperform local features.
At around 25\% remaining points, a subsampled KITTI frame has roughly the same number of points as nuScenes, indicating that global location data should be preferred when a detector is pre-trained on sparse data and local neighborhoods are unreliable.
However, this scenario is unlikely in real-world applications.
Notably, as sparsity increases, the combined model ($xyz$ + GBlobs) degrades less significantly than GBlobs alone, suggesting it leverages GBlobs in dense voxels while relying on global coordinates elsewhere.

\section{Conclusion}
\label{sec:conclusion}

In this work, we emphasized the critical role of local point cloud geometry in building robust LiDAR-based 3D detection models.
Our evaluations revealed that relying solely on standard global input features leads to poor cross-domain generalization.
To address this limitation, we introduced a highly efficient, parameter-free approach that can be seamlessly integrated into any 3D object detector: encoding local point cloud geometry using Gaussian blobs.
Our method, \ourmethod, has been extensively validated across a range of detectors and benchmark scenarios, demonstrating competitive in-domain performance (on par with standard global features) while achieving significant improvements in cross-domain generalization.

\nbf{Acknowledgments} We gratefully acknowledge the financial support by the Austrian Federal Ministry for Digital and Economic Affairs, the National Foundation for Research, Technology and Development and the Christian Doppler Research Association. The presented experiments have been achieved using the Vienna Scientific Cluster.

{\small
\bibliographystyle{ieeenat_fullname}
\bibliography{abbrv_short,11_references}

\begin{thebibliography}{70}
\providecommand{\natexlab}[1]{#1}
\providecommand{\url}[1]{\texttt{#1}}
\expandafter\ifx\csname urlstyle\endcsname\relax
  \providecommand{\doi}[1]{doi: #1}\else
  \providecommand{\doi}{doi: \begingroup \urlstyle{rm}\Url}\fi

\bibitem[Addepalli et~al.(2024)Addepalli, Asokan, Sharma, and
  Babu]{addepalli2024leveraging}
Sravanti Addepalli, Ashish~Ramayee Asokan, Lakshay Sharma, and R~Venkatesh
  Babu.
\newblock {Leveraging Vision-Language Models for Improving Domain
  Generalization in Image Classification}.
\newblock In \emph{Proc. CVPR}, 2024.

\bibitem[Caesar et~al.(2020)Caesar, Bankiti, Lang, Vora, Liong, Xu, Krishnan,
  Pan, Baldan, and Beijbom]{caesar2020nuscenes}
Holger Caesar, Varun Bankiti, Alex~H Lang, Sourabh Vora, Venice~Erin Liong,
  Qiang Xu, Anush Krishnan, Yu Pan, Giancarlo Baldan, and Oscar Beijbom.
\newblock {nuScenes: A multimodal dataset for autonomous driving}.
\newblock In \emph{Proc. CVPR}, 2020.

\bibitem[Caine et~al.(2021)Caine, Roelofs, Vasudevan, Ngiam, Chai, Chen, and
  Shlens]{Caine2021}
Benjamin Caine, Rebecca Roelofs, Vijay Vasudevan, Jiquan Ngiam, Yuning Chai,
  Zhifeng Chen, and Jonathon Shlens.
\newblock {P}seudo-labeling for {S}calable {3D} {O}bject {D}etection.
\newblock \emph{arXiv CoRR}, abs/2103.02093, 2021.

\bibitem[Chen et~al.(2023{\natexlab{a}})Chen, Liu, Zhang, Qi, and
  Jia]{chen2023voxelnext}
Yukang Chen, Jianhui Liu, Xiangyu Zhang, Xiaojuan Qi, and Jiaya Jia.
\newblock {VoxelNeXt: Fully Sparse VoxelNet for 3D Object Detection and
  Tracking}.
\newblock In \emph{Proc. CVPR}, 2023{\natexlab{a}}.

\bibitem[Chen et~al.(2023{\natexlab{b}})Chen, Luo, Wang, Baktashmotlagh, and
  Huang]{chen2023revisiting}
Zhuoxiao Chen, Yadan Luo, Zheng Wang, Mahsa Baktashmotlagh, and Zi Huang.
\newblock {Revisiting Domain-Adaptive 3D Object Detection by Reliable, Diverse
  and Class-balanced Pseudo-Labeling}.
\newblock In \emph{Proc. ICCV}, 2023{\natexlab{b}}.

\bibitem[Chen et~al.(2024)Chen, Wang, Zhao, Su, Men, and
  Meng]{chen2024practicaldg}
Zining Chen, Weiqiu Wang, Zhicheng Zhao, Fei Su, Aidong Men, and Hongying Meng.
\newblock {PracticalDG: Perturbation Distillation on Vision-Language Models for
  Hybrid Domain Generalization}.
\newblock In \emph{Proc. CVPR}, 2024.

\bibitem[Choi et~al.(2021)Choi, Song, and Kwak]{choi2021part}
Jaeseok Choi, Yeji Song, and Nojun Kwak.
\newblock {Part-Aware Data Augmentation for 3D Object Detection in Point
  Cloud}.
\newblock In \emph{Proc. IROS}, 2021.

\bibitem[Deng et~al.(2021)Deng, Shi, Li, Zhou, Zhang, and Li]{deng2021voxel}
Jiajun Deng, Shaoshuai Shi, Peiwei Li, Wengang Zhou, Yanyong Zhang, and
  Houqiang Li.
\newblock {Voxel R-CNN: Towards High Performance Voxel-based 3D Object
  Detection}.
\newblock In \emph{Proc. AAAI}, 2021.

\bibitem[Eskandar(2024)]{eskandar2024empirical}
George Eskandar.
\newblock {An Empirical Study of the Generalization Ability of Lidar 3D Object
  Detectors to Unseen Domains}.
\newblock In \emph{Proc. CVPR}, 2024.

\bibitem[Fan et~al.(2022)Fan, Pang, Zhang, Wang, Zhao, Wang, Wang, and
  Zhang]{fan2022embracing}
Lue Fan, Ziqi Pang, Tianyuan Zhang, Yu-Xiong Wang, Hang Zhao, Feng Wang, Naiyan
  Wang, and Zhaoxiang Zhang.
\newblock {Embracing Single Stride 3D Object Detector with Sparse Transformer}.
\newblock In \emph{Proc. CVPR}, 2022.

\bibitem[Fruhwirth-Reisinger et~al.(2021)Fruhwirth-Reisinger, Opitz, Possegger,
  and Bischof]{FruhwirthReisinger2021}
Christian Fruhwirth-Reisinger, Michael Opitz, Horst Possegger, and Horst
  Bischof.
\newblock {FAST3D}: {F}low-{A}ware {S}elf-{T}raining for {3D} {O}bject
  {D}etectors.
\newblock In \emph{Proc. BMVC}, 2021.

\bibitem[Geiger et~al.(2012)Geiger, Lenz, and Urtasun]{geiger2012kitti}
Andreas Geiger, Philip Lenz, and Raquel Urtasun.
\newblock {Are we ready for Autonomous Driving? The KITTI Vision Benchmark
  Suite}.
\newblock In \emph{Proc. CVPR}, 2012.

\bibitem[Gong et~al.(2019)Gong, Li, Chen, and Gool]{gong2019dlow}
Rui Gong, Wen Li, Yuhua Chen, and Luc~Van Gool.
\newblock {DLOW: Domain Flow for Adaptation and Generalization}.
\newblock In \emph{Proc. CVPR}, 2019.

\bibitem[Hahner et~al.(2020)Hahner, Dai, Liniger, and Van~Gool]{hahner2020aug}
Martin Hahner, Dengxin Dai, Alexander Liniger, and Luc Van~Gool.
\newblock {Quantifying Data Augmentation for LiDAR-based 3D Object Detection}.
\newblock \emph{arXiv CoRR}, abs/2004.01643, 2020.

\bibitem[Hoppe et~al.(1992)Hoppe, DeRose, Duchamp, McDonald, and
  Stuetzle]{hoppe1992surface}
Hugues Hoppe, Tony DeRose, Tom Duchamp, John McDonald, and Werner Stuetzle.
\newblock {Surface Reconstruction from Unorganized Points}.
\newblock In \emph{SIGGRAPH}, 1992.

\bibitem[Hu et~al.(2023)Hu, Liu, and Hu]{hu2023density}
Qianjiang Hu, Daizong Liu, and Wei Hu.
\newblock {Density-Insensitive Unsupervised Domain Adaption on 3D Object
  Detection}.
\newblock In \emph{Proc. CVPR}, 2023.

\bibitem[Huang et~al.(2021)Huang, Guan, Xiao, and Lu]{huang2021fsdr}
Jiaxing Huang, Dayan Guan, Aoran Xiao, and Shijian Lu.
\newblock {FSDR: Frequency Space Domain Randomization for Domain
  Generalization}.
\newblock In \emph{Proc. CVPR}, 2021.

\bibitem[Kim et~al.(2024)Kim, Woo, and Im]{jaeyeul2024eccv}
Jaeyeul Kim, Jeonghoon Woo, Jungwan sand~Kim, and Sunghoon Im.
\newblock {Rethinking LiDAR Domain Generalization: Single Source as Multiple
  Density Domains}.
\newblock In \emph{Proc. ECCV}, 2024.

\bibitem[Kolesnikov et~al.(2021)Kolesnikov, Dosovitskiy, Weissenborn, Heigold,
  Uszkoreit, Beyer, Minderer, Dehghani, Houlsby, Gelly, Unterthiner, and
  Zhai]{alexey2020image}
Alexander Kolesnikov, Alexey Dosovitskiy, Dirk Weissenborn, Georg Heigold,
  Jakob Uszkoreit, Lucas Beyer, Matthias Minderer, Mostafa Dehghani, Neil
  Houlsby, Sylvain Gelly, Thomas Unterthiner, and Xiaohua Zhai.
\newblock {An Image is Worth 16x16 Words: Transformers for Image Recognition at
  Scale}.
\newblock In \emph{Proc. ICML}, 2021.

\bibitem[Lang et~al.(2019)Lang, Vora, Caesar, Zhou, Yang, and
  Beijbom]{lang2019pointpillars}
Alex~H Lang, Sourabh Vora, Holger Caesar, Lubing Zhou, Jiong Yang, and Oscar
  Beijbom.
\newblock {PointPillars: Fast Encoders for Object Detection from Point Clouds}.
\newblock In \emph{Proc. CVPR}, 2019.

\bibitem[Lehner et~al.(2022)Lehner, Gasperini, Marcos-Ramiro, Schmidt, Mahani,
  Navab, Busam, and Tombari]{lehner20223d}
Alexander Lehner, Stefano Gasperini, Alvaro Marcos-Ramiro, Michael Schmidt,
  Mohammad-Ali~Nikouei Mahani, Nassir Navab, Benjamin Busam, and Federico
  Tombari.
\newblock {3D-VField: Adversarial Augmentation of Point Clouds for Domain
  Generalization in 3D Object Detection}.
\newblock In \emph{Proc. CVPR}, 2022.

\bibitem[Leng et~al.(2023)Leng, Li, Liu, Cubuk, Sun, He, Anguelov, and
  Tan]{leng2023lidar}
Zhaoqi Leng, Guowang Li, Chenxi Liu, Ekin~Dogus Cubuk, Pei Sun, Tong He,
  Dragomir Anguelov, and Mingxing Tan.
\newblock {LidarAugment: Searching for Scalable 3D LiDAR Data Augmentations}.
\newblock In \emph{Proc. ICRA}, 2023.

\bibitem[Li et~al.(2018{\natexlab{a}})Li, Pan, Wang, and Kot]{li2018cvprdg}
Haoliang Li, Sinno~Jialin Pan, Shiqi Wang, and Alex~C Kot.
\newblock {Domain Generalization with Adversarial Feature Learning}.
\newblock In \emph{Proc. CVPR}, 2018{\natexlab{a}}.

\bibitem[Li et~al.(2023{\natexlab{a}})Li, Luo, and Yang]{li2023pillarnext}
Jinyu Li, Chenxu Luo, and Xiaodong Yang.
\newblock {PillarNeXt: Rethinking network designs for 3D object detection in
  LiDAR point clouds}.
\newblock In \emph{Proc. CVPR}, 2023{\natexlab{a}}.

\bibitem[Li et~al.(2024{\natexlab{a}})Li, Qiao, Shum, and Breckon]{li2024trail}
Li Li, Tanqiu Qiao, Hubert~PH Shum, and Toby~P Breckon.
\newblock {TraIL-Det: Transformation-Invariant Local Feature Networks for 3D
  LiDAR Object Detection with Unsupervised Pre-Training}.
\newblock In \emph{Proc. BMVC}, 2024{\natexlab{a}}.

\bibitem[Li et~al.(2024{\natexlab{b}})Li, Shum, and Breckon]{li2024rapid}
Li Li, Hubert~PH Shum, and Toby~P Breckon.
\newblock {RAPiD-Seg: Range-Aware Pointwise Distance Distribution Networks for
  3D LiDAR Segmentation}.
\newblock In \emph{Proc. ECCV}, 2024{\natexlab{b}}.

\bibitem[Li et~al.(2023{\natexlab{b}})Li, Feng, Shi, Gao, Fang, Liu, and
  Han]{li2023neuralgf}
Qing Li, Huifang Feng, Kanle Shi, Yue Gao, Yi Fang, Yu-Shen Liu, and Zhizhong
  Han.
\newblock {NeuralGF: Unsupervised Point Normal Estimation by Learning Neural
  Gradient Function}.
\newblock In \emph{Proc. NeurIPS}, 2023{\natexlab{b}}.

\bibitem[Li et~al.(2024{\natexlab{c}})Li, Ma, and Li]{li2023domainne2021}
Shuangzhi Li, Lei Ma, and Xingyu Li.
\newblock {Domain Generalization of 3D Object Detection by Density-Resampling}.
\newblock In \emph{Proc. ECCV}, 2024{\natexlab{c}}.

\bibitem[Li et~al.(2018{\natexlab{b}})Li, Gong, Tian, Liu, and
  Tao]{li2018aaaidg}
Ya Li, Mingming Gong, Xinmei Tian, Tongliang Liu, and Dacheng Tao.
\newblock Domain generalization via conditional invariant representations.
\newblock In \emph{Proc. AAAI}, 2018{\natexlab{b}}.

\bibitem[Liu et~al.(2021)Liu, Lin, Cao, Hu, Wei, Zhang, Lin, and
  Guo]{liu2021swin}
Ze Liu, Yutong Lin, Yue Cao, Han Hu, Yixuan Wei, Zheng Zhang, Stephen Lin, and
  Baining Guo.
\newblock {Swin Transformer: Hierarchical Vision Transformer using Shifted
  Windows}.
\newblock In \emph{Proc. ICCV}, 2021.

\bibitem[Liu et~al.(2022)Liu, Mao, Wu, Feichtenhofer, Darrell, and
  Xie]{liu2022convnet}
Zhuang Liu, Hanzi Mao, Chao-Yuan Wu, Christoph Feichtenhofer, Trevor Darrell,
  and Saining Xie.
\newblock {A ConvNet for the 2020s}.
\newblock In \emph{Proc. CVPR}, 2022.

\bibitem[Lv et~al.(2022)Lv, Liang, Li, Zang, Liu, Wang, and
  Liu]{lv2022causality}
Fangrui Lv, Jian Liang, Shuang Li, Bin Zang, Chi~Harold Liu, Ziteng Wang, and
  Di Liu.
\newblock {Causality Inspired Representation Learning for Domain
  Generalization}.
\newblock In \emph{Proc. CVPR}, 2022.

\bibitem[Mahajan et~al.(2021)Mahajan, Tople, and Sharma]{mahajan2021domain}
Divyat Mahajan, Shruti Tople, and Amit Sharma.
\newblock {Domain Generalization using Causal Matching}.
\newblock In \emph{Proc. ICML}, 2021.

\bibitem[Mali{\'c} et~al.(2023)Mali{\'c}, Fruhwirth-Reisinger, Possegger, and
  Bischof]{malic2023sailor}
Du{\v{s}}an Mali{\'c}, Christian Fruhwirth-Reisinger, Horst Possegger, and
  Horst Bischof.
\newblock {SAILOR: Scaling Anchors via Insights into Latent Object
  Representation}.
\newblock In \emph{Proc. WACV}, 2023.

\bibitem[Mao et~al.(2022)Mao, Niu, Jiang, Liang, Chen, Liang, Li, Ye, Zhang,
  Li, et~al.]{mao2022one}
Jiageng Mao, Minzhe Niu, Chenhan Jiang, Hanxue Liang, Jingheng Chen, Xiaodan
  Liang, Yamin Li, Chaoqiang Ye, Wei Zhang, Zhenguo Li, et~al.
\newblock {One Million Scenes for Autonomous Driving: ONCE Dataset}.
\newblock \emph{Proc. NeurIPS}, 2022.

\bibitem[Meyer et~al.(2019)Meyer, Laddha, Kee, Vallespi-Gonzalez, and
  Wellington]{meyer2019lasernet}
Gregory~P Meyer, Ankit Laddha, Eric Kee, Carlos Vallespi-Gonzalez, and Carl~K
  Wellington.
\newblock {LaserNet: An Efficient Probabilistic 3D Object Detector for
  Autonomous Driving}.
\newblock In \emph{Proc. CVPR}, 2019.

\bibitem[Miao et~al.(2021)Miao, Hirakawa, Yamashita, and Fujiyoshi]{miao20213d}
Jishu Miao, Tsubasa Hirakawa, Takayoshi Yamashita, and Hironobu Fujiyoshi.
\newblock {3D Object Detection with Normal-map on Point Clouds.}
\newblock In \emph{Proc. VISIGRAPP}, 2021.

\bibitem[Naich and Carri{\'o}n(2024)]{naich2024lidar}
Ammar~Yasir Naich and Jes{\'u}s~Requena Carri{\'o}n.
\newblock {LiDAR-Based Intensity-Aware Outdoor 3D Object Detection}.
\newblock \emph{Sensors}, 24\penalty0 (9):\penalty0 2942, 2024.

\bibitem[Peng et~al.(2019)Peng, Huang, Sun, and Saenko]{peng2019domain}
Xingchao Peng, Zijun Huang, Ximeng Sun, and Kate Saenko.
\newblock {Domain Agnostic Learning with Disentangled Representations}.
\newblock In \emph{Proc. ICML}, 2019.

\bibitem[Qi et~al.(2017)Qi, Su, Mo, and Guibas]{Qi_2017_CVPR}
Charles~R. Qi, Hao Su, Kaichun Mo, and Leonidas~J. Guibas.
\newblock {PointNet: Deep Learning on Point Sets for 3D Classification and
  Segmentation}.
\newblock In \emph{Proc. CVPR}, 2017.

\bibitem[Qi et~al.(2018)Qi, Liu, Wu, Su, and Guibas]{qi2018frustum}
Charles~R. Qi, Wei Liu, Chenxia Wu, Hao Su, and Leonidas~J Guibas.
\newblock {Frustum PointNets for 3D Object Detection from RGB-D Data}.
\newblock In \emph{Proc. CVPR}, 2018.

\bibitem[Rame et~al.(2022)Rame, Dancette, and Cord]{rame2022fishr}
Alexandre Rame, Corentin Dancette, and Matthieu Cord.
\newblock {Fishr: Invariant Gradient Variances for Out-of-Distribution
  Generalization}.
\newblock In \emph{Proc. ICML}, 2022.

\bibitem[Ran et~al.(2022)Ran, Liu, and Wang]{ran2022surface}
Haoxi Ran, Jun Liu, and Chengjie Wang.
\newblock {Surface Representation for Point Clouds}.
\newblock In \emph{Proc. CVPR}, 2022.

\bibitem[Rist et~al.(2019)Rist, Enzweiler, and Gavrila]{rist2019cross}
Christoph~B Rist, Markus Enzweiler, and Dariu~M Gavrila.
\newblock {Cross-Sensor Deep Domain Adaptation for LiDAR Detection and
  Segmentation}.
\newblock In \emph{Proc. IV}, 2019.

\bibitem[Scheuble et~al.(2024)Scheuble, Lei, Baek, Bijelic, and
  Heide]{scheuble2024polarization}
Dominik Scheuble, Chenyang Lei, Seung-Hwan Baek, Mario Bijelic, and Felix
  Heide.
\newblock {Polarization Wavefront Lidar: Learning Large Scene Reconstruction
  from Polarized Wavefronts}.
\newblock In \emph{Proc. CVPR}, 2024.

\bibitem[Shi et~al.(2019)Shi, Wang, and Li]{shi2019pointrcnn}
Shaoshuai Shi, Xiaogang Wang, and Hongsheng Li.
\newblock {PointRCNN: 3D Object Proposal Generation and Detection from Point
  Cloud}.
\newblock In \emph{Proc. CVPR}, 2019.

\bibitem[Shi et~al.(2020)Shi, Wang, Shi, Wang, and Li]{shi2020points}
Shaoshuai Shi, Zhe Wang, Jianping Shi, Xiaogang Wang, and Hongsheng Li.
\newblock {From Points to Parts: 3D Object Detection From Point Cloud With
  Part-Aware and Part-Aggregation Network}.
\newblock \emph{TPAMI}, 43\penalty0 (8):\penalty0 2647--2664, 2020.

\bibitem[Shi et~al.(2023)Shi, Jiang, Deng, Wang, Guo, Shi, Wang, and
  Li]{shi2023pv}
Shaoshuai Shi, Li Jiang, Jiajun Deng, Zhe Wang, Chaoxu Guo, Jianping Shi,
  Xiaogang Wang, and Hongsheng Li.
\newblock {PV-RCNN++: Point-Voxel Feature Set Abstraction With Local Vector
  Representation for 3D Object Detection}.
\newblock \emph{IJCV}, 131\penalty0 (2):\penalty0 531--551, 2023.

\bibitem[Soum-Fontez et~al.(2023)Soum-Fontez, Deschaud, and
  Goulette]{soum2023mdt3d}
Louis Soum-Fontez, Jean-Emmanuel Deschaud, and Fran{\c{c}}ois Goulette.
\newblock {MDT3D: Multi-Dataset Training for LiDAR 3D Object Detection
  Generalization}.
\newblock In \emph{Proc. IROS}, 2023.

\bibitem[Sun et~al.(2020)Sun, Kretzschmar, Dotiwalla, Chouard, Patnaik, Tsui,
  Guo, Zhou, Chai, Caine, et~al.]{sun2020wod}
Pei Sun, Henrik Kretzschmar, Xerxes Dotiwalla, Aurelien Chouard, Vijaysai
  Patnaik, Paul Tsui, James Guo, Yin Zhou, Yuning Chai, Benjamin Caine, et~al.
\newblock {Scalability in Perception for Autonomous Driving: Waymo Open
  Dataset}.
\newblock In \emph{Proc. CVPR}, 2020.

\bibitem[Sun et~al.(2022)Sun, Tan, Wang, Liu, Xia, Leng, and
  Anguelov]{sun2022swformer}
Pei Sun, Mingxing Tan, Weiyue Wang, Chenxi Liu, Fei Xia, Zhaoqi Leng, and
  Dragomir Anguelov.
\newblock {SWFormer: Sparse Window Transformer for 3D Object Detection in Point
  Clouds}.
\newblock In \emph{Proc. ECCV}, 2022.

\bibitem[Vaswani et~al.(2017)Vaswani, Shazeer, Parmar, Uszkoreit, Jones, Gomez,
  Kaiser, and Polosukhin]{vaswani2017attention}
Ashish Vaswani, Noam Shazeer, Niki Parmar, Jakob Uszkoreit, Llion Jones,
  Aidan~N. Gomez, \L{}ukasz Kaiser, and Illia Polosukhin.
\newblock {Attention Is All You Need}.
\newblock In \emph{Proc. NeurIPS}, 2017.

\bibitem[Volpi and Murino(2019)]{volpi2019addressing}
Riccardo Volpi and Vittorio Murino.
\newblock {Addressing Model Vulnerability to Distributional Shifts over Image
  Transformation Sets}.
\newblock In \emph{Proc. ICCV}, 2019.

\bibitem[Walsh et~al.(2020)Walsh, Ku, Pon, and Waslander]{Walsh2020}
Sean Walsh, Jason Ku, Alex~D. Pon, and Steven~L. Waslander.
\newblock {L}everaging {T}emporal {D}ata for {A}utomatic {L}abelling of
  {S}tatic {V}ehicles.
\newblock In \emph{Proc. CVPR}, 2020.

\bibitem[Wang et~al.(2023)Wang, Shi, Shi, Lei, Wang, He, Schiele, and
  Wang]{wang2023dsvt}
Haiyang Wang, Chen Shi, Shaoshuai Shi, Meng Lei, Sen Wang, Di He, Bernt
  Schiele, and Liwei Wang.
\newblock {DSVT: Dynamic Sparse Voxel Transformer with Rotated Sets}.
\newblock In \emph{Proc. CVPR}, 2023.

\bibitem[Wang et~al.(2020)Wang, Chen, You, Li, Hariharan, Campbell, Weinberger,
  and Chao]{wang2020trainingermany}
Yan Wang, Xiangyu Chen, Yurong You, Li~Erran Li, Bharath Hariharan, Mark
  Campbell, Kilian~Q Weinberger, and Wei-Lun Chao.
\newblock {Train in Germany, Test in The USA: Making 3D Object Detectors
  Generalize}.
\newblock In \emph{Proc. CVPR}, 2020.

\bibitem[Wang et~al.(2024)Wang, Han, Liu, and Li]{wang2024ride}
Zhaoxuan Wang, Xu Han, Hongxin Liu, and Xianzhi Li.
\newblock {RIDE: Boosting 3D Object Detection for LiDAR Point Clouds via
  Rotation-Invariant Analysis}.
\newblock \emph{arXiv CoRR}, abs/2408.15643, 2024.

\bibitem[Wei et~al.(2022)Wei, Wei, Rao, Li, Zhou, and Lu]{wei2022lidar}
Yi Wei, Zibu Wei, Yongming Rao, Jiaxin Li, Jie Zhou, and Jiwen Lu.
\newblock {LiDAR Distillation: Bridging the Beam-Induced Domain Gap for 3D
  Object Detection}.
\newblock In \emph{Proc. ECCV}, 2022.

\bibitem[Wu et~al.(2023)Wu, Cao, Liu, Chen, and Ren]{wu2023towards}
Guile Wu, Tongtong Cao, Bingbing Liu, Xingxin Chen, and Yuan Ren.
\newblock {Towards Universal LiDAR-Based 3D Object Detection by Multi-Domain
  Knowledge Transfer}.
\newblock In \emph{Proc. ICCV}, 2023.

\bibitem[Wu et~al.(2024)Wu, Jiang, Wang, Liu, Liu, Qiao, Ouyang, He, and
  Zhao]{wu2024ptv3}
Xiaoyang Wu, Li Jiang, Peng-Shuai Wang, Zhijian Liu, Xihui Liu, Yu Qiao, Wanli
  Ouyang, Tong He, and Hengshuang Zhao.
\newblock {Point Transformer V3: Simpler, Faster, Stronger}.
\newblock In \emph{Proc. CVPR}, 2024.

\bibitem[Yan et~al.(2018)Yan, Mao, and Li]{yan2018second}
Yan Yan, Yuxing Mao, and Bo Li.
\newblock {SECOND: Sparsely Embedded Convolutional Detection}.
\newblock \emph{Sensors}, 18\penalty0 (10):\penalty0 3337, 2018.

\bibitem[Yang et~al.(2021)Yang, Shi, Wang, Li, and Qi]{3dda:st3d}
Jihan Yang, Shaoshuai Shi, Zhe Wang, Hongsheng Li, and Xiaojuan Qi.
\newblock {ST3D}: {S}elf-training for {U}nsupervised {D}omain {A}daptation on
  {3D} {O}bject {D}etection.
\newblock In \emph{Proc. CVPR}, 2021.

\bibitem[Yang et~al.(2022)Yang, Shi, Wang, Li, and Qi]{yang2022st3d++}
Jihan Yang, Shaoshuai Shi, Zhe Wang, Hongsheng Li, and Xiaojuan Qi.
\newblock {ST3D++: Denoised Self-Training for Unsupervised Domain Adaptation on
  3D Object Detection}.
\newblock \emph{TPAMI}, 45\penalty0 (5):\penalty0 6354--6371, 2022.

\bibitem[Yang et~al.(2020)Yang, Sun, Liu, and Jia]{yang20203dssd}
Zetong Yang, Yanan Sun, Shu Liu, and Jiaya Jia.
\newblock {3DSSD: Point-based 3D Single Stage Object Detector}.
\newblock In \emph{Proc. CVPR}, 2020.

\bibitem[Yin et~al.(2021)Yin, Zhou, and Krahenbuhl]{yin2021center}
Tianwei Yin, Xingyi Zhou, and Philipp Krahenbuhl.
\newblock {Center-based 3D Object Detection and Tracking}.
\newblock In \emph{Proc. CVPR}, 2021.

\bibitem[You et~al.(2022)You, Diaz-Ruiz, Wang, Chao, Hariharan, Campbell, and
  Weinbergert]{you2022exploiting}
Yurong You, Carlos~Andres Diaz-Ruiz, Yan Wang, Wei-Lun Chao, Bharath Hariharan,
  Mark Campbell, and Kilian~Q Weinbergert.
\newblock {E}xploiting {P}laybacks in {U}nsupervised {D}omain {A}daptation for
  {3D} {O}bject {D}etection.
\newblock In \emph{Proc. ICRA}, 2022.

\bibitem[Yue et~al.(2019)Yue, Zhang, Zhao, Sangiovanni-Vincentelli, Keutzer,
  and Gong]{yue2019domain}
Xiangyu Yue, Yang Zhang, Sicheng Zhao, Alberto Sangiovanni-Vincentelli, Kurt
  Keutzer, and Boqing Gong.
\newblock {Domain Randomization and Pyramid Consistency: Simulation-to-Real
  Generalization without Accessing Target Domain Data}.
\newblock In \emph{Proc. ICCV}, 2019.

\bibitem[Zhang et~al.(2024)Zhang, Chen, Xiao, Peng, Li, Lin, Li, Wang, Wu, and
  Cai]{zhang2024pseudo}
Zhanwei Zhang, Minghao Chen, Shuai Xiao, Liang Peng, Hengjia Li, Binbin Lin,
  Ping Li, Wenxiao Wang, Boxi Wu, and Deng Cai.
\newblock {Pseudo Label Refinery for Unsupervised Domain Adaptation on
  Cross-dataset 3D Object Detection}.
\newblock In \emph{Proc. CVPR}, 2024.

\bibitem[Zhao et~al.(2021)Zhao, Zhu, Liang, and Chen]{zhao2021integration}
Yan Zhao, Jihong Zhu, Haoyu Liang, and Lyujie Chen.
\newblock {Integration of Coordinate and Geometric Surface Normal for 3D Point
  Cloud Object Detection}.
\newblock In \emph{Proc. IJCNN}, 2021.

\bibitem[Zhou and Tuzel(2018)]{zhou2018voxelnet}
Yin Zhou and Oncel Tuzel.
\newblock {VoxelNet: End-to-End Learning for Point Cloud Based 3D Object
  Detection}.
\newblock In \emph{Proc. CVPR}, 2018.

\end{thebibliography}


\begin{thebibliography}{12}
\providecommand{\natexlab}[1]{#1}
\providecommand{\url}[1]{\texttt{#1}}
\expandafter\ifx\csname urlstyle\endcsname\relax
  \providecommand{\doi}[1]{doi: #1}\else
  \providecommand{\doi}{doi: \begingroup \urlstyle{rm}\Url}\fi

\bibitem[Caesar et~al.(2020)Caesar, Bankiti, Lang, Vora, Liong, Xu, Krishnan,
  Pan, Baldan, and Beijbom]{caesar2020nuscenes}
Holger Caesar, Varun Bankiti, Alex~H Lang, Sourabh Vora, Venice~Erin Liong,
  Qiang Xu, Anush Krishnan, Yu Pan, Giancarlo Baldan, and Oscar Beijbom.
\newblock {nuScenes: A multimodal dataset for autonomous driving}.
\newblock In \emph{Proc. CVPR}, 2020.

\bibitem[Choi et~al.(2021)Choi, Song, and Kwak]{choi2021part}
Jaeseok Choi, Yeji Song, and Nojun Kwak.
\newblock {Part-Aware Data Augmentation for 3D Object Detection in Point
  Cloud}.
\newblock In \emph{Proc. IROS}, 2021.

\bibitem[Deng et~al.(2021)Deng, Shi, Li, Zhou, Zhang, and Li]{deng2021voxel}
Jiajun Deng, Shaoshuai Shi, Peiwei Li, Wengang Zhou, Yanyong Zhang, and
  Houqiang Li.
\newblock {Voxel R-CNN: Towards High Performance Voxel-based 3D Object
  Detection}.
\newblock In \emph{Proc. AAAI}, 2021.

\bibitem[Geiger et~al.(2012)Geiger, Lenz, and Urtasun]{geiger2012kitti}
Andreas Geiger, Philip Lenz, and Raquel Urtasun.
\newblock {Are we ready for Autonomous Driving? The KITTI Vision Benchmark
  Suite}.
\newblock In \emph{Proc. CVPR}, 2012.

\bibitem[Lang et~al.(2019)Lang, Vora, Caesar, Zhou, Yang, and
  Beijbom]{lang2019pointpillars}
Alex~H Lang, Sourabh Vora, Holger Caesar, Lubing Zhou, Jiong Yang, and Oscar
  Beijbom.
\newblock {PointPillars: Fast Encoders for Object Detection from Point Clouds}.
\newblock In \emph{Proc. CVPR}, 2019.

\bibitem[Lehner et~al.(2022)Lehner, Gasperini, Marcos-Ramiro, Schmidt, Mahani,
  Navab, Busam, and Tombari]{lehner20223d}
Alexander Lehner, Stefano Gasperini, Alvaro Marcos-Ramiro, Michael Schmidt,
  Mohammad-Ali~Nikouei Mahani, Nassir Navab, Benjamin Busam, and Federico
  Tombari.
\newblock {3D-VField: Adversarial Augmentation of Point Clouds for Domain
  Generalization in 3D Object Detection}.
\newblock In \emph{Proc. CVPR}, 2022.

\bibitem[Sun et~al.(2020)Sun, Kretzschmar, Dotiwalla, Chouard, Patnaik, Tsui,
  Guo, Zhou, Chai, Caine, et~al.]{sun2020wod}
Pei Sun, Henrik Kretzschmar, Xerxes Dotiwalla, Aurelien Chouard, Vijaysai
  Patnaik, Paul Tsui, James Guo, Yin Zhou, Yuning Chai, Benjamin Caine, et~al.
\newblock {Scalability in Perception for Autonomous Driving: Waymo Open
  Dataset}.
\newblock In \emph{Proc. CVPR}, 2020.

\bibitem[Wang et~al.(2023)Wang, Shi, Shi, Lei, Wang, He, Schiele, and
  Wang]{wang2023dsvt}
Haiyang Wang, Chen Shi, Shaoshuai Shi, Meng Lei, Sen Wang, Di He, Bernt
  Schiele, and Liwei Wang.
\newblock {DSVT: Dynamic Sparse Voxel Transformer with Rotated Sets}.
\newblock In \emph{Proc. CVPR}, 2023.

\bibitem[Yan et~al.(2018)Yan, Mao, and Li]{yan2018second}
Yan Yan, Yuxing Mao, and Bo Li.
\newblock {SECOND: Sparsely Embedded Convolutional Detection}.
\newblock \emph{Sensors}, 18\penalty0 (10):\penalty0 3337, 2018.

\bibitem[Yang et~al.(2021)Yang, Shi, Wang, Li, and Qi]{3dda:st3d}
Jihan Yang, Shaoshuai Shi, Zhe Wang, Hongsheng Li, and Xiaojuan Qi.
\newblock {ST3D}: {S}elf-training for {U}nsupervised {D}omain {A}daptation on
  {3D} {O}bject {D}etection.
\newblock In \emph{Proc. CVPR}, 2021.

\bibitem[Yang et~al.(2022)Yang, Shi, Wang, Li, and Qi]{yang2022st3d++}
Jihan Yang, Shaoshuai Shi, Zhe Wang, Hongsheng Li, and Xiaojuan Qi.
\newblock {ST3D++: Denoised Self-Training for Unsupervised Domain Adaptation on
  3D Object Detection}.
\newblock \emph{TPAMI}, 45\penalty0 (5):\penalty0 6354--6371, 2022.

\bibitem[Yin et~al.(2021)Yin, Zhou, and Krahenbuhl]{yin2021center}
Tianwei Yin, Xingyi Zhou, and Philipp Krahenbuhl.
\newblock {Center-based 3D Object Detection and Tracking}.
\newblock In \emph{Proc. CVPR}, 2021.

\end{thebibliography}
}

\ifarxiv \clearpage \appendix \maketitlesupplementary This supplementary material details the experimental setup (\cref{sec:experimental_setup}), presents further findings on the influence of local and global features on LiDAR-based 3D object detection (\cref{sec:height_bias}), and provides a qualitative analysis of our results (\cref{sec:qualitative_res}).

\section{Experimental Setup}
\label{sec:experimental_setup}

To ensure reproducibility (besides the provided source code), we present detailed information about our experimental setup in \cref{tab:detailed_experiments}.
If not specified otherwise, we use default settings from OpenPCDet\footnote{\url{https://github.com/open-mmlab/OpenPCDet/}}.

Our models were trained on the entire KITTI and nuScenes training sets, along with $20\%$ of the Waymo dataset (a standard practice in the field).
In all our experiments (except \KtoW in Tab. 3 of the main manuscript), we trained the models to simultaneously predict Cars/Vehicles, Pedestrians, and Cyclists.
For fair comparison with 3D-VF~\cite{lehner20223d}, we trained the detector to predict only Cars/Vehicles in \KtoW.
We employed standard data augmentation techniques, including random sampling, point cloud rotation, scaling, and flipping.

We use the KITTI metric~\cite{geiger2012kitti} for evaluation (except in Tab. 6b of the main manuscript), reporting Average Precision (AP) on Bird's-eye View (BEV) / 3D views at $40$ recall positions.
For the in-domain Waymo evaluation in Tab. 6b (main manuscript), we report LEVEL\_1 /LEVEL\_2 AP (standard Waymo metric).
We use Intersection over Union (IoU) thresholds of $0.7$, $0.5$, and $0.5$ for Cars, Pedestrians, and Cyclists, respectively.
\KtoW in Tab. 3 (main manuscript) uses an IoU threshold of $0.5$ for Cars to ensure fair comparison.
We utilize the complete validation sets of all datasets to assess the performance of our proposed method.

\begin{table*}
\scriptsize
\centering
\resizebox{\textwidth}{!}{%
\begin{tabular}{lllccccccc}
\toprule
\multirow{4}{*}{Table}  & \multirow{4}{*}{Dataset}           & \multicolumn{6}{c}{Detector}             \\
\cmidrule{3-8}
                        &                                    & \multirow{2}{*}{Name}                     & \multirow{2}{*}{Range}                                                                    & \multirow{2}{*}{Voxel Size}              & \multicolumn{3}{c}{Optimizer} & \multirow{1}{*}{BS} & \multirow{1}{*}{E} \\
\cmidrule{6-8}
                        &                                    &                                           &                                                                                           &                                          & Name                          & LR                  & WD                  &      \\
\midrule
\multirow{2}{*}{Tab. 2} & Waymo~\cite{sun2020wod}            & Voxel R-CNN~\cite{deng2021voxel}          & $[-75.2\phantom{0}, -75.2\phantom{0}, -2, 75.2\phantom{0}, 75.2\phantom{0}, 4]$           & $[0.1\phantom{0}, 0.1\phantom{0}, 0.15]$ & Adam                          & $1 \times 10^{-2}$  & $1 \times 10^{-3}$  & $32$  & $30$ \\
                        & nuScenes~\cite{caesar2020nuscenes} & Voxel R-CNN~\cite{deng2021voxel}          & $[-75.2\phantom{0}, -75.2\phantom{0}, -2, 75.2\phantom{0}, 75.2\phantom{0}, 4]$           & $[0.1\phantom{0}, 0.1\phantom{0}, 0.15]$ & Adam                          & $1 \times 10^{-2}$  & $1 \times 10^{-3}$  & $32$  & $30$ \\
\midrule
\multirow{3}{*}{Tab. 3} & KITTI~\cite{geiger2012kitti}       & PointPillars~\cite{lang2019pointpillars}  & $[\phantom{-0}0.0\phantom{0}, -39.68, -2, 69.12, 39.68, 4]$                               & $[0.16, 0.16, 6.0\phantom{0}]$           & Adam                          & $3 \times 10^{-3}$  & $1 \times 10^{-2}$  & $32$  & $80$ \\
                        & KITTI~\cite{geiger2012kitti}       & SECOND~\cite{yan2018second}               & $[\phantom{-0}0.0\phantom{0}, -40.0\phantom{0}, -3, 70.4\phantom{0}, 40.0\phantom{0}, 1]$ & $[0.16, 0.16, 6.0\phantom{0}]$           & Adam                          & $3 \times 10^{-3}$  & $1 \times 10^{-2}$  & $32$  & $80$ \\
                        & KITTI~\cite{geiger2012kitti}       & Part-A$^2$~\cite{choi2021part}            & $[\phantom{-0}0.0\phantom{0}, -40.0\phantom{0}, -3, 70.4\phantom{0}, 40.0\phantom{0}, 1]$ & $[0.16, 0.16, 6.0\phantom{0}]$           & Adam                          & $1 \times 10^{-2}$  & $1 \times 10^{-2}$  & $32$  & $80$ \\
\midrule
Tab. 4                  & $*$                                & CenterPoint~\cite{yin2021center}          & $[-75.2\phantom{0}, -75.2\phantom{0}, -3, 75.2\phantom{0}, 75.2\phantom{0}, 5]$           & $[0.10, 0.10, 0.20]$                     & Adam                          & $3 \times 10^{-3}$  & $1 \times 10^{-2}$  & $32$  & $30$ \\
\midrule
Tab. 5                  & nuScenes~\cite{caesar2020nuscenes} & Voxel R-CNN~\cite{deng2021voxel}          & $[-75.2\phantom{0}, -75.2\phantom{0}, -2, 75.2\phantom{0}, 75.2\phantom{0}, 4]$           & $[0.1\phantom{0}, 0.1\phantom{0}, 0.15]$ & Adam                          & $1 \times 10^{-2}$  & $1 \times 10^{-3}$  & $32$  & $30$ \\
\midrule
Tab. 6                  & KITTI~\cite{geiger2012kitti}       & SECOND~\cite{yan2018second}               & $[\phantom{-0}0.0\phantom{0}, -40.0\phantom{0}, -3, 70.4\phantom{0}, 40.0\phantom{0}, 1]$ & $[0.16, 0.16, 6.0\phantom{0}]$           & Adam                          & $3 \times 10^{-3}$  & $1 \times 10^{-2}$  & $32$  & $80$ \\
Tab. 6a                 & Waymo~\cite{sun2020wod}            & DSVT~\cite{wang2023dsvt}                  & $[-74.88, -74.88, -2, 74.88, 74.88, 4]$                                                   & $[0.32, 0.32, 6.0\phantom{0}]$           & Adam                          & $3 \times 10^{-3}$  & $5 \times 10^{-2}$  & $24$  & $30$ \\
Tab. 6b                 & Waymo~\cite{sun2020wod}            & Voxel R-CNN~\cite{deng2021voxel}          & $[-75.2\phantom{0}, -75.2\phantom{0}, -2, 75.2\phantom{0}, 75.2\phantom{0}, 4]$           & $[0.1\phantom{0}, 0.1\phantom{0}, 0.15]$ & Adam                          & $1 \times 10^{-2}$  & $1 \times 10^{-3}$  & $32$  & $30$ \\
\bottomrule
\end{tabular}
}%
\caption{
  Complete experimental setup for each table from the main manuscript.
  We specify the source domain, where $*$ specifies all except the target dataset for our multi-source domain generalization.
  We report LiDAR point cloud range, voxel size, optimizer parameters (learning rate (LR), weight decay (WD)), batch size (BS) and the number of epochs (E) used for training.
}
\label{tab:detailed_experiments}
\end{table*}

\section{Height Bias}
\label{sec:height_bias}

Autonomous driving datasets define different reference points for LiDAR point clouds, \eg
Waymo~\cite{sun2020wod} aligns the height axis origin with the road, while KITTI~\cite{geiger2012kitti} and nuScenes~\cite{caesar2020nuscenes} use the vehicle's mounting point.
This inherently introduces bias into the network.
A common approach is to manually align source and target point clouds by shifting them to a shared origin~\cite{yang2022st3d++,3dda:st3d}.
Otherwise the detectors fail catastrophically as demonstrated in \cref{tab:z_shift}.
Although this is not a critical issue in our controlled setting, a detector trained with such bias could pose a significant risk in real-world applications.
Our \ourmethod are not affected by biases associated with global input features, as they encodes local point cloud geometry.

\begin{figure}
  \centering
  \includegraphics[width=0.8\linewidth]{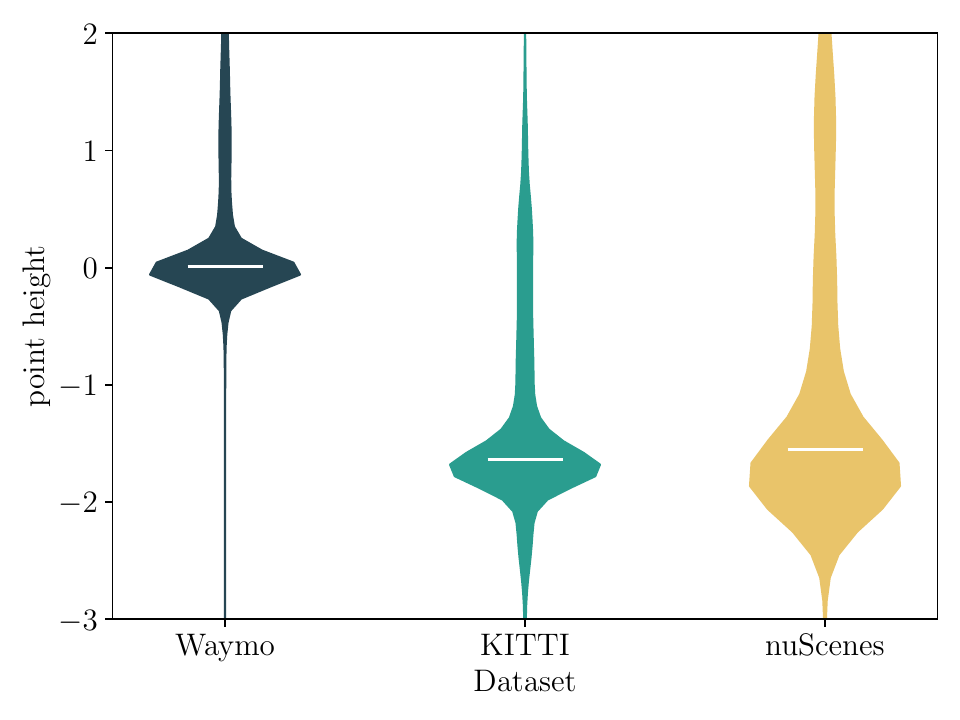}
  \caption{$z$ distribution}
  \label{fig:z distribution}
\end{figure}

\begin{table*}
\centering
\begin{tabular}{clcccc}
\toprule
\textbf{$z$-alignment}      & \textbf{Method}                                & \textbf{Car}                          & \textbf{Pedestrian}                   & \textbf{Cyclist}                      & \textbf{mAP}                          \\
\midrule
\multirow{2}{*}{\checkmark} & Voxel R-CNN~\cite{deng2021voxel}               & 66.93/28.80                           & 23.39/18.65                           & 19.23/15.76                           & 36.52/21.07                           \\
                            & Voxel R-CNN~\cite{deng2021voxel} w/ \ourmethod & \bestresult{80.95}/\bestresult{53.98} & \bestresult{38.33}/\bestresult{33.22} & \bestresult{29.18}/\bestresult{25.68} & \bestresult{49.48}/\bestresult{37.62} \\
\midrule
\multirow{2}{*}{\ding{55}}  & Voxel R-CNN~\cite{deng2021voxel}               & {54.61}/20.83                         & 10.51/\phantom{0}7.68                 & \phantom{0}5.88/\phantom{0}5.12       & 23.66/11.21                           \\
                            & Voxel R-CNN~\cite{deng2021voxel} w/ \ourmethod & \bestresult{80.84}/\bestresult{55.05} & \bestresult{37.93}/\bestresult{33.24} & \bestresult{28.62}/\bestresult{24.60} & \bestresult{49.13}/\bestresult{37.63} \\
\bottomrule
\end{tabular}
\caption{
  Influence of $z$-alignment on detector trained with different input features.
  We trained Voxel R-CNN~\cite{deng2021voxel} on nuScenes~\cite{caesar2020nuscenes} using all three classes (Car, Pedestrian, Cyclist) simultaneously and evaluated performance using Average Precision (AP) on Bird's-eye View (BEV) / 3D views at $40$ recall positions.
  Intersection over Union (IoU) thresholds of $0.7$, $0.5$, and $0.5$ were used for Car, Pedestrian, and Cyclist, respectively.
  We evaluate the performance on KITTI~\cite{geiger2012kitti}, where we report the average AP across all difficulty levels (Easy, Moderate, Hard).
  Additionally, we provide the mean AP over the three classes.
  The best value in each category is highlighted in bold.
}
\label{tab:z_shift}
\end{table*}

\section{Qualitative Results}
\label{sec:qualitative_res}

In order to depict benefits of training a model with our \ourmethod as input features, we conduct following qualitative analysis.
We apply a nuScenes trained Voxel R-CNN detector to a challenging KITTI scene featuring a slightly curved road.
Such detectors, trained with standard global input features, often predict false positives, even in areas without object indications.

A similar phenomenon can be observed with the SECOND~\cite{yan2018second} detector employed in the \KtoW benchmark in \cref{fig:quantitative_analysis_k2w}.
It is noteworthy that SECOND, trained on KITTI, a dataset consisting primarily of small and mid-size European sedans, has never seen anything that resembles aerial work platforms during training.
Nevertheless, when trained with global features and applied on Waymo (which has such object labeled as Vehicles), it manages to produce a detection at this location with high certainty (orange arrows in \cref{fig:quantitative_analysis_global_k2w}).
We hypothesize that the detector's prediction was influenced by specific points at specific heights.
Given its training, such detections are unexpected.
This raises the question of how many other detections, which are false positives, such detector produces.
A model trained with our \ourmethod did not make such uneducated guess (\cref{fig:quantitative_analysis_ours_k2w}).

\begin{figure*}
     \centering
     \begin{subfigure}{\textwidth}
         \centering
         \includegraphics[width=\textwidth]{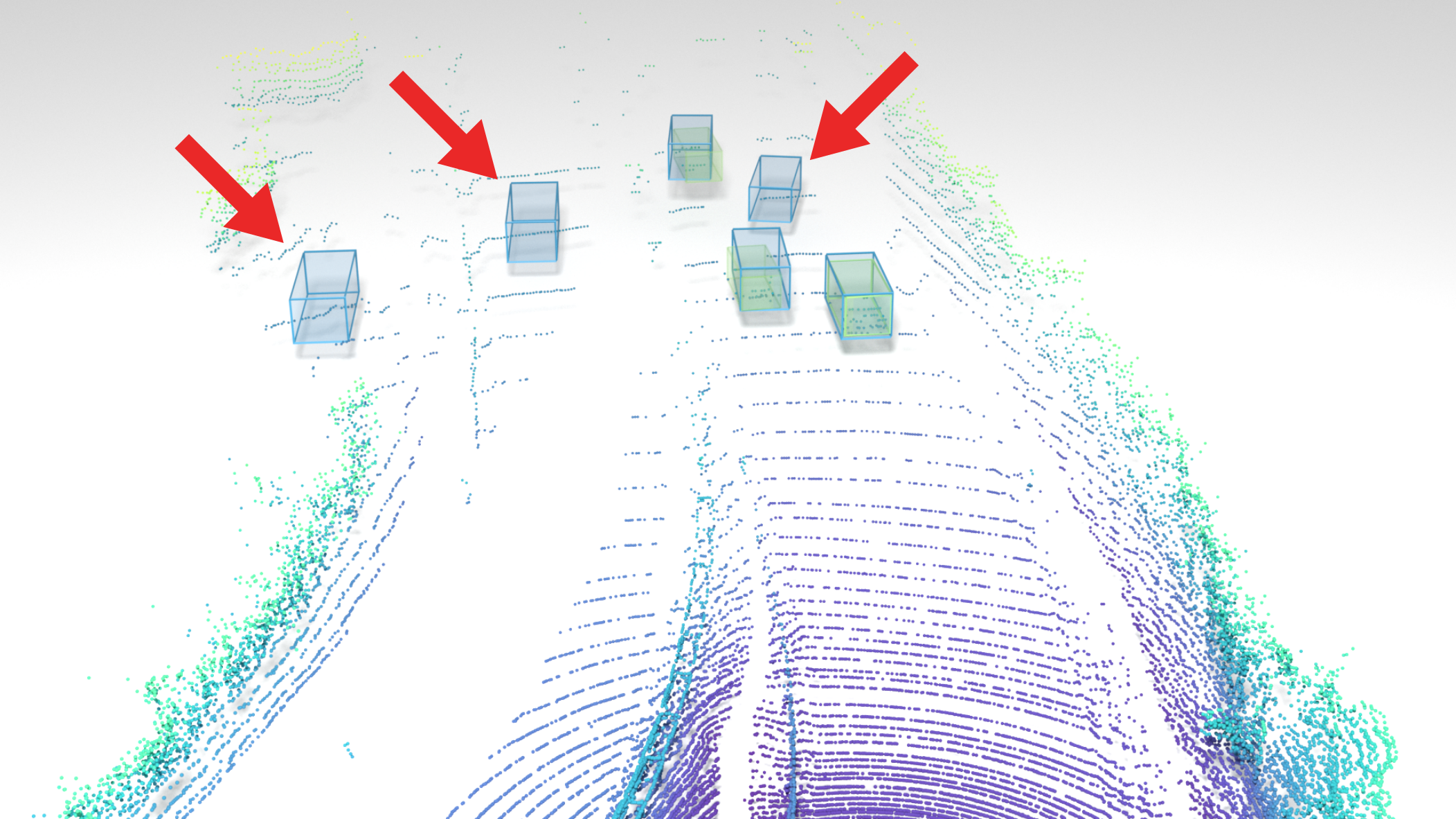}
         \caption{
           \NtoK Voxel R-CNN~\cite{deng2021voxel}.
         }
         \label{fig:quantitative_analysis_global}
     \end{subfigure} \hfill
     \begin{subfigure}{\textwidth}
         \centering
         \includegraphics[width=\textwidth]{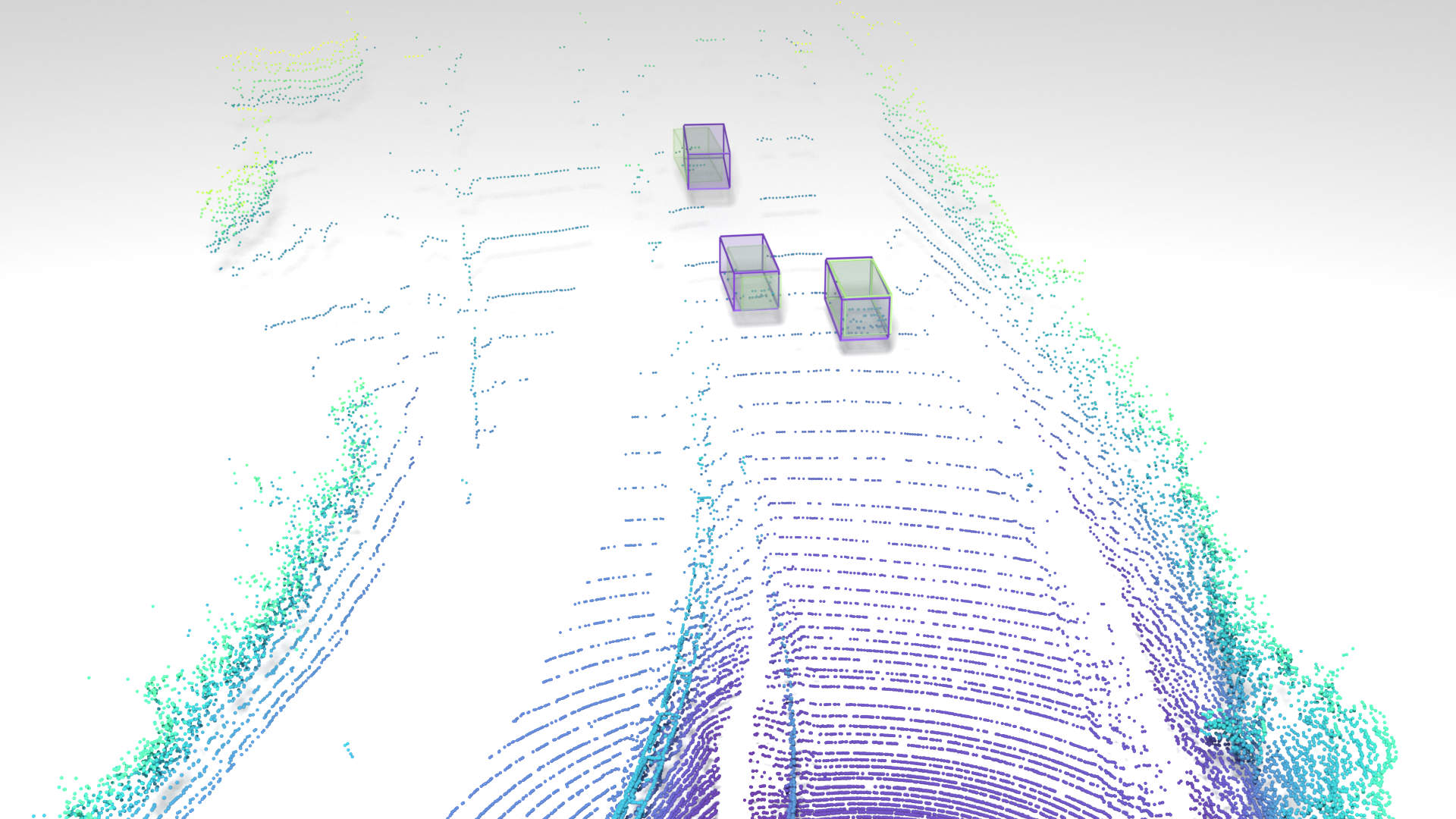}
         \caption{
           \NtoK Voxel R-CNN~\cite{deng2021voxel} w/ \ourmethod.
         }
         \label{fig:quantitative_analysis_ours}
     \end{subfigure}
     \caption{
       Qualitative evaluation of Voxel R-CNN~\cite{deng2021voxel} on a \NtoK benchmark thresholded at $0.5$.
       Ground truth detections are shown in green.
       Detections from a model trained on standard global input features and our \ourmethod are depicted in blue (a) and purple (b), respectively.
       False positive detections are marked with red arrows.
       The color of the point cloud represents the height.
     }
     \label{fig:quantitative_analysis}
\end{figure*}

\begin{figure*}
     \centering
     \begin{subfigure}{\textwidth}
         \centering
         \includegraphics[width=\textwidth]{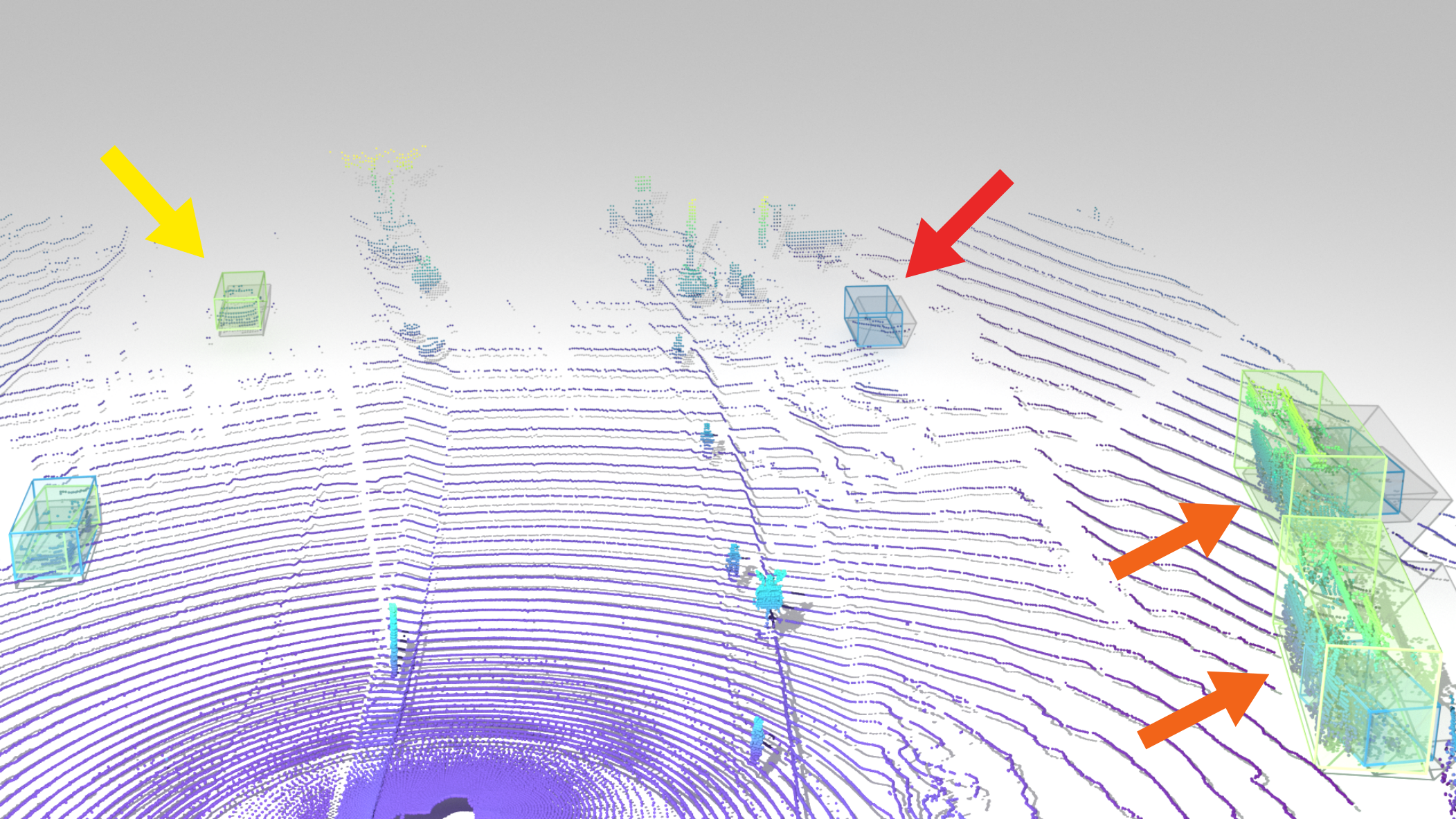}
         \caption{
           \KtoW SECOND~\cite{yan2018second}.
         }
         \label{fig:quantitative_analysis_global_k2w}
     \end{subfigure} \hfill
     \begin{subfigure}{\textwidth}
         \centering
         \includegraphics[width=\textwidth]{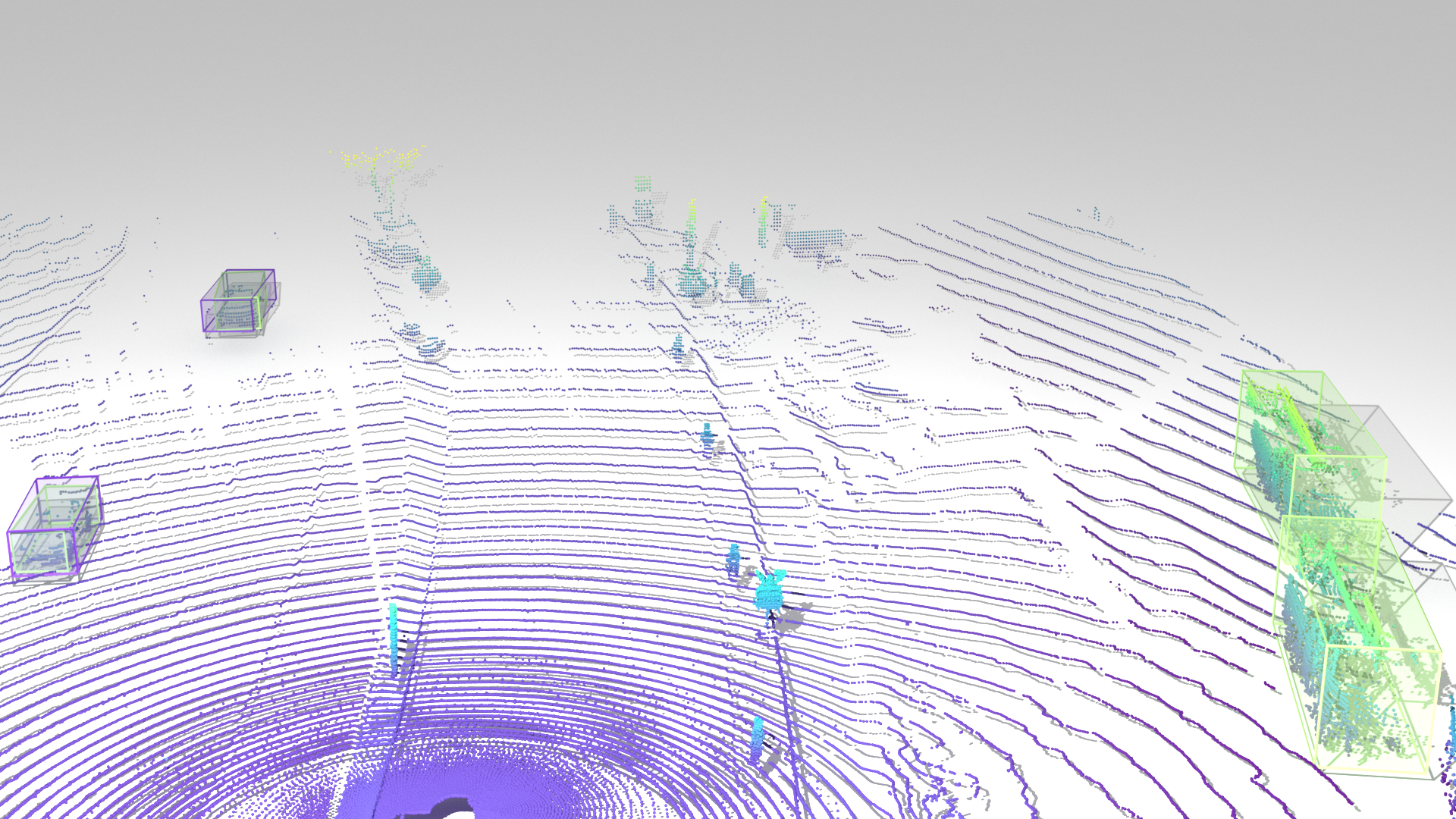}
         \caption{
           \KtoW SECOND~\cite{yan2018second} w/ \ourmethod.
         }
         \label{fig:quantitative_analysis_ours_k2w}
     \end{subfigure}
     \caption{
       Qualitative evaluation of SECOND~\cite{yan2018second} on a \KtoW benchmark thresholded at $0.5$.
       Ground truth detections are shown in green.
       Detections from a model trained on standard global input features and our \ourmethod are depicted in blue (a) and purple (b), respectively.
       False positive and false negative detections are marked with red and yellow arrow, respetively.
       Detections which are dubious are markes with orange arrow.
       The color of the point cloud represents the height.
     }
     \label{fig:quantitative_analysis_k2w}
\end{figure*}

 \fi

\end{document}